\documentclass{article} % For LaTeX2e
\usepackage{iclr2026_conference,times}
\usepackage{amsmath,amsfonts}
\usepackage{algorithmic}
\usepackage{algorithm}
\usepackage{array}
\usepackage[caption=false,font=normalsize]{subfig}
\usepackage{textcomp}
\usepackage{stfloats}
\usepackage{url}
\usepackage{verbatim}
\usepackage{graphicx}
\usepackage{subfig}
\usepackage{cite}
\usepackage{booktabs} 
\usepackage{multirow}
\usepackage{adjustbox}
\usepackage{afterpage}
\usepackage{soul, color, xcolor}

\usepackage[utf8]{inputenc}
\usepackage{CJKutf8}

\usepackage{caption}

\usepackage{hyperref}

\title{Beyond English-Centric Training: How Reinforcement Learning Improves Cross-Lingual Reasoning in LLMs}

\author{
 \textbf{Shulin Huang$^{1,2}$\thanks{~ indicates equal contribution.}},
 \textbf{Yiran Ding$^{2*}$},
 \textbf{Junshu Pan$^{1,2*}$},
    \textbf{Yue Zhang$^{2}$\thanks{~Correspondence to: (zhangyue@westlake.edu.cn)}} \\
 \textsuperscript{1}Zhejiang University,
 \textsuperscript{2}Westlake University\\
\texttt{\{huangshulin,dingyiran,panjunshu,zhangyue\}@westlake.edu.cn} \\
}

\iclrfinalcopy % Uncomment for camera-ready version, but NOT for submission.
\begin{document}

\maketitle

\begin{abstract}
% Enhancing the reasoning capabilities of large language models (LLMs) across diverse languages is a critical yet challenging frontier in AI. 
% Enhancing the complex reasoning capabilities of Large Language Models (LLMs) represents a key frontier.
Enhancing the complex reasoning capabilities of Large Language Models (LLMs) attracts widespread attention.
While reinforcement learning (RL) has shown superior performance for improving complex reasoning, its impact on cross-lingual generalization compared to Supervised Fine-Tuning (SFT) remains unexplored. We present the first systematic investigation into cross-lingual reasoning generalization of RL and SFT. Using Qwen2.5-3B-Base as our foundation model, we conduct experiments on diverse multilingual reasoning benchmarks, including math reasoning, commonsense reasoning, and scientific reasoning. Our investigation yields two significant findings: (1) Tuning with RL not only achieves higher accuracy but also demonstrates substantially stronger cross-lingual generalization capabilities compared to SFT. (2) RL training on non-English data yields better overall performance and generalization than training on English data, which is not observed with SFT. 
% This finding challenges the conventional wisdom of English-centric model training.
Furthermore, through comprehensive mechanistic analyses, we explore the underlying factors of RL's superiority and generalization across languages.
Our results provide compelling evidence that RL enables the model with more robust reasoning strategies, offering crucial guidance for more equitable and effective multilingual reasoning.
\end{abstract}

\section{Introduction}

\begin{figure}[h]
    \centering
    \includegraphics[width=0.98\textwidth]{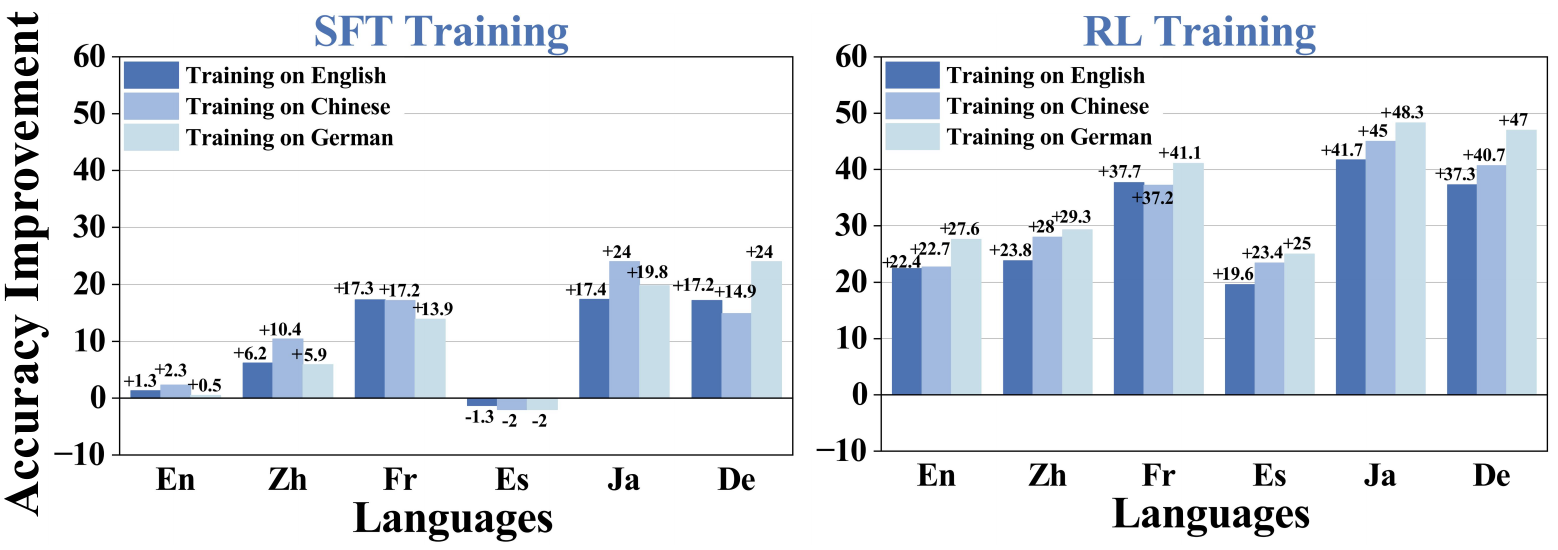}
    % %\vspace{-10pt}
    \caption{SFT and RL performance improvement using English, Chinese and German training data on the same base model, Qwen2.5-3B-Base. Performance improvements are measured relative to the base model. We report the performance improvement in six language settings.
    % Some evaluations extend their assessment to include English language evaluation at most. 
    %Within the limited computation budget, the selected process reward model used for test-time compute model usually should consider the trade-off between the efficiency and the accuracy performance.
    }
    \label{fig:intro}
    %\vspace{-15pt}
\end{figure}

Multilingualism plays a significant role in human society and occupies a critical position in the development of large language models (LLMs). With over 7,000 languages worldwide, each encapsulates unique cultural contexts and expressive modalities~\citep{campbell2008ethnologue}. LLMs not only break down language barriers and facilitate cross-cultural communication, but also enable equitable global Artificial Intelligence benefits~\citep{howard2018universal,sharma2025faux}.
% LLMs, which are capable of serving users across different linguistic communities, not only break down language barriers and facilitate cross-cultural communication, but also enable equitable access to artificial intelligence benefits for users globally~\citep{howard2018universal,sharma2025faux}.

% In multilingual artificial intelligence systems, cross-lingual generalization of reasoning capabilities represents a crucial research direction. 
% As research progresses beyond basic multilingual comprehension, increasing attention is being directed toward Multilingual reasoning.
With advances in LLMs reasoning~\citep{hao2023reasoning,yue2025does}, cross-lingual reasoning has received increasing attention~\citep{wang2024seaeval,chai2025xcot,payoungkhamdee2025towards}.
Multilingual reasoning requires models to not only comprehend the semantic content of different languages, but also to possess the ability to perform logical inference and problem-solving across diverse linguistic environments~\citep{alam2024llms}. Current research demonstrates that while large-scale pre-trained language models have achieved remarkable progress in English comprehension and generation, significant performance gaps persist for other languages~\citep{qiu2024cross}. 
% This disparity manifests not only in substantial variations in reasoning accuracy across different languages, but also in the depth of understanding that models demonstrate regarding language-specific reasoning patterns. 
Consequently, enhancing multilingual reasoning capabilities and achieving effective cross-lingual generalization has emerged as a significant challenge~\citep{pham2024unibridge,hu2025large}.

Reinforcement Learning (RL) is considered a pivotal tool for enhancing reasoning capabilities~\citep{wang2024reinforcement,guo2025deepseek}. Recent studies show that RL training exhibits superior performance on complex tasks such as mathematical and logical reasoning~\citep{xie2025logic}. Compared to Supervised Fine-tuning (SFT), Reinforcement Learning, through reward-guided mechanisms, enables models with more robust and generalizable reasoning strategies~\citep{huan2025does}. Notably, researches reveal that RL not only significantly improves model performance, but also enables stronger \textit{cross-task} generalization capabilities~\citep{shao2024deepseekmath}. 
% This advantage becomes particularly pronounced when handling complex problems requiring multi-step reasoning, where models can progressively optimize their reasoning paths through trial-and-error learning, thereby achieving higher problem-solving accuracy [11].
In this work, we conduct the first investigation into whether RL exhibits strong generalization capabilities across languages in reasoning. Experimentally, we compare the performance of RL and SFT on diverse languages, exploring their performance across various languages to examine their \textit{cross-lingual} generalization abilities. 
To fully reflect reasoning capability, we evaluate performance on diverse multilingual reasoning
benchmarks, including math reasoning, commonsense reasoning, and scientific
reasoning~\citep{shilanguage,she2024mapo,son2025linguistic,xuan2025mmlu,qi2025models}.

% \section{Related Work}

The empirical investigation yields two notable findings: (1) As illustrated in Figure~\ref{fig:intro}, RL demonstrates superior performance improvements compared to SFT, with enhanced cross-lingual generalization capabilities. Our results indicate that models trained with RL can more effectively transfer reasoning abilities learned in one language to another. This finding of cross-lingual reasoning is consistent with existing findings for cross-task transfer~\citep{korkmaz2024survey,huan2025does,cheng2025revisiting,chusft}. (2) 
Given that the pre-training corpora of most existing LLMs are predominantly English-centric~\citep{morishita2024enhancing,rytting2021leveraging,singh2024global}, the conventional expectation is that RL training with English data would maximally leverage the model's potential~\citep{yoon2024langbridge,she2024mapo}. However, as shown in Figure 1, our findings surprisingly reveal a counter-intuitive phenomenon: RL training using non-English data (such as Chinese and German) yields better cross-lingual reasoning performance and superior generalization than using English data.
% As shown in Figure~\ref{fig:intro}, surprisingly, RL training using non-English data yields better cross-lingual reasoning performance and superior generalization compared to using English data for RL training. 
% This phenomenon is not observed with supervised fine-tuning data, challenging the conventional works of English as the preferred training language and suggesting that different languages may offer unique advantages.
This contrasts with SFT, where no such phenomenon is observed, as performance remains comparable and, in some datasets, even shows an opposite trend.
% These findings challenge the conventional view~\citep{yoon2024langbridge,she2024mapo} of English as the preferred training language and suggest that different languages may offer unique advantages.

% 为了探究这两个发现背后的深层原因，我们设计了两组实验。首先，我们通过分析模型在推理过程中“思考”所使用的语言与问题语言是否一致，来探究语言转换现象对跨语言泛化能力的影响。其次，我们引入了拒绝采样微调（Rejection Sampling Fine-Tuning, RFT）作为对比，深入剖析了强化学习中的“探索-优化”采样过程相较于监督微调的优势所在。这些实验使我们能够从语言一致性和训练范式两个维度，系统地解释强化学习为何能取得更优越的跨语言推理表现。

% To investigate the underlying reasons for these two findings, we conduct in-depth mechanistic analyses. 
% First, we analyze whether the language used by the model for its thinking was consistent with the language of the question, to explore the impact of this linguistic inconsistency on cross-lingual generalization. Second, we introduce Rejection Sampling Fine-Tuning (RFT) as a comparative baseline to dissect the advantages of the ``explore-and-optimize'' sampling process in RL over SFT. 
% These two mechanisms collectively explain why RL can surpass SFT and achieve superior cross-lingual generalization capabilities.
% These experiments allow us to systematically explain why RL achieves superior cross-lingual reasoning performance from the perspectives of language consistency and training paradigms.

% 为了理解强化学习为何在跨语言推理方面表现出如此显著的优势，我们进行了深入的机制分析。通过对模型生成的推理过程进行详细检查，我们发现了两个关键因素：首先，强化学习训练的模型倾向于在推理过程中使用与问题语言不一致的混合语言或通用语言，这种语言不一致性使模型能够利用更强大的语言无关推理模块；其次，强化学习的采样机制通过探索多样化的解题路径并基于奖励信号进行策略优化，使模型学习到更加鲁棒和通用的问题解决方法。这两个机制共同解释了为什么强化学习能够超越监督微调并实现卓越的跨语言泛化能力。
To investigate underlying reasons for the findings, we conduct preliminary analyses: (1) First, we analyze whether the language used in reasoning is consistent with the language of input question. Our investigation reveals that language inconsistency serves as a potential factor of RL's cross-lingual generalization, and the superiority of the non-English data in RL. (2) Second, we examine the role of sampling mechanisms in RL's superior performance. We find that the sampling mechanism in RL explores sufficient and diverse solution paths, allowing models to learn more robust and generalizable strategies. (3) Third, we explore the semantic shift of the model after training. We find that the stability of the semantic space contributes to RL's superior cross-lingual generalization. Our preliminary explorations provide insights for future research in multilingual reasoning.

The main contributions of this work are as follows: 

(1) We present the first systematic analysis of the differences between RL and SFT in cross-lingual reasoning generalization, filling an important gap in this research area.

(2) We reveal two significant findings: 1) RL excels over SFT in cross-lingual generalization, and 2)
Counterintuitively, non-English data is superior to English data for RL training. 
To our knowledge, we are the first to demonstrate that using non-English data for RL more effectively enhances performance and cross-lingual generalization, although most models are pre-trained mainly on English.
% To our knowledge, we are the first to demonstrate that even models pre-trained mainly on English, RL with non-English data more effectively enhances performance and cross-lingual generalization.
% We reveal two significant findings: RL not only delivers superior performance compared to SFT but also significantly enhances cross-lingual reasoning generalization. Surprisingly, RL training with non-English data yields better performance and generalization than with English data. To our knowledge, we are the first to identify this counterintuitive phenomenon: despite large language models being predominantly pre-trained on English data, RL training with non-English data yields superior performance and cross-lingual generalization than English-based training.

(3) Through comprehensive analyses, we explore three potential factors underlying RL's superiority: linguistic inconsistency in reasoning, sampling-driven policy optimization, and the semantic shift after training, which provides a crucial foundation for multilingual reasoning.
%
% (2) Based on in-depth analyses, we reveal that reinforcement learning, compared to supervised fine-tuning, not only delivers superior performance but also significantly enhances cross-lingual reasoning generalization capabilities, providing crucial theoretical guidance and practical recommendations for multilingual reasoning.
%
%
%\vspace{-10pt}
\section{Related Work}
%\vspace{-5pt}
% \paragraph{Multilingual Reasoning.}
\textbf{Multilingual Reasoning.}
% 多语言数学推理是评估大语言模型智能水平的重要任务（Ahn et al., 2024）。Shi et al.（2022）将GSM8K数据集（Cobbe et al., 2021）中的英文数学题翻译成多种语言，构建了多语言数学推理基准MGSM，为该领域奠定了基础。针对提升多语言推理能力，现有工作主要采用提示工程方法：Qin et al.（2023）和Huang et al.（2023）提出先将非英语问题翻译成英语再求解的策略，在ChatGPT等闭源模型上取得良好效果。然而，这些方法在开源模型上的有效性尚未得到充分验证，为开源大语言模型赋予强大的多语言数学推理能力仍是一个开放性挑战。此外，Yuan et al.（2023）和Chen et al.（2024）进一步探索了跨语言知识迁移和代码辅助推理等方法来增强多语言推理性能
Multilingual reasoning is a challenging and representative task for evaluating the intelligence of large language models~\citep{ahn2024large,she2024mapo,yoon2024langbridge}. ~\citet{shilanguage} establish the foundation for this field by translating English mathematical problems from GSM8K~\citep{cobbe2021gsm8k} into multiple languages, creating the multilingual benchmark MGSM~\citep{shilanguage}. To enhance multilingual reasoning capabilities, existing work primarily employs prompting strategies.~\citet{qin2023cross} and ~\citet{huang2023not} propose a translate-then-solve approach that first translates non-English questions into English before problem-solving, achieving promising results on closed-source models like ChatGPT~\citep{ouyang2022training}. 
% However, the effectiveness of these methods on open-source models remains insufficiently validated, and equipping open-source LLMs with strong multilingual mathematical reasoning capabilities remains an open challenge. 
% Additionally, Yuan et al. (2023) and Chen et al. (2024) further explored cross-lingual knowledge transfer and code-assisted reasoning to enhance multilingual reasoning performance.
% 然而，先前的工作主要集中在通过提示工程（prompting strategies）提升性能，而很少关注不同训练范式（如SFT与RL）对模型内在跨语言泛化能力的影响。本文则另辟蹊径，不依赖于翻译或特定的提示技巧，而是系统地比较了SFT和RL在模型底层能力上的差异，揭示了RL在学习语言无关（language-agnostic）推理策略上的独特优势。
However, less attention has been paid to how different training paradigms affect the model's intrinsic cross-lingual generalization capabilities. Our work addresses this gap by comparing SFT and RL at the model's foundational level, demonstrating RL's unique advantage in learning reasoning strategies without relying on specific languages.
% However, there is a lack of attention paid to how different training paradigms (such as SFT vs. RL) affect the model's intrinsic cross-lingual generalization capabilities. Our work takes a different approach; instead of relying on translation or specific prompting techniques, we systematically compare the differences between SFT and RL at the model's foundational level, revealing RL's unique advantage in learning language-agnostic reasoning strategies.

% \paragraph{Supervised Fine-tuning For Reasoning.}
\textbf{Supervised Fine-tuning For Reasoning.}
% 监督微调（SFT）通过蒸馏专家级思维链（CoT）样本有效提升了大语言模型的推理能力。合成数据生成是主流策略：Mammoth（Yue et al., 2023）、MathScale（Tang et al., 2024）和WizardMath（Luo et al., 2023b）利用大型教师模型通过自指导方法（Wang et al., 2022a）为数学问题生成解答。OpenMathInstruct-2（Toshniwal et al., 2024）和MetaMath（Yu et al., 2023）进一步研究了数据质量因素的影响。除数学领域外，LLaVA-CoT-100k（Xu et al., 2024）和Kim et al.（Kim et al., 2023）的CoT集合将推理问题扩展到更大规模。近期工作如PRIME（Korbak et al., 2024）和Self-Taught Reasoner（Zelikman et al., 2022）还探索了迭代自训练和自我一致性等技术。
Supervised fine-tuning (SFT) effectively enhances LLM reasoning abilities by distilling expert-level chain-of-thought (CoT) examples~\citep{huang2024o1}. Synthetic data generation is a key strategy: Large teacher models are used to generate solutions for mathematical problems~\citep{yue2023mammoth,tang2024mathscale}, enhancing the reasoning process. Additionally, recent research examines the impact of data quality factors on the model performance ~\citep{toshniwal2024openmathinstruct2,yu2023metamath,ye2025limo}. Beyond mathematics, other works~\citep{kim2023cot,xu2024llavacot100k} expand reasoning tasks to larger domains, broadening the scope and complexity of problem-solving in various fields.
% Supervised fine-tuning (SFT) effectively enhances LLM reasoning by distilling expert-level chain-of-thought (CoT) examples~\citep{huang2024o1}. Synthetic data generation is a key strategy, with models like Mammoth~\citep{yue2023mammoth}, MathScale~\citep{tang2024mathscale}, and WizardMath~\citep{luo2023wizardmath} using large teacher models for problem-solving through self-instruct methods~\citep{wang2022selfinstruct}. OpenMathInstruct-2~\citep{toshniwal2024openmathinstruct2}, MetaMath~\citep{yu2023metamath}, and LIMO~\citep{ye2025limo} explore data quality factors. Other works~\citep{kim2023cot,xu2024llavacot100k} extend reasoning tasks to broader domains, expanding problem-solving capabilities.
% 尽管SFT在增强特定任务的推理能力上取得了成功，但其学习方式本质上是模仿给定的“专家”轨迹，这可能导致模型对训练数据的语言和模式产生过拟合。我们的研究与这些工作不同，我们着重探讨了SFT在跨语言场景下的局限性，并将其与RL进行对比，证明了仅仅依赖高质量的CoT数据进行模仿，难以实现像RL那样强大的跨语言泛化。
Although SFT is successful in enhancing reasoning, its learning approach is fundamentally based on imitating and memorizing given ``expert'' trajectories~\citep{ge2023supervised}, leading to overfitting to the language and pattern in the training data. Our research differs from these works by focusing on the limitations of SFT in cross-lingual scenarios. By contrasting it with RL, we demonstrate that merely imitating high-quality CoT data is insufficient for achieving the robust cross-lingual generalization that RL provides.

% \paragraph{Reinforcement Learning For Reasoning.}
\textbf{Reinforcement Learning For Reasoning.}
% 强化学习（RL）广泛应用于大语言模型的后训练阶段以对齐人类偏好（Ouyang et al., 2022; Grattafiori et al., 2024），近期研究将其扩展至推理能力增强。这些方法将CoT推理建模为强化学习问题，采用多样化奖励机制，包括最终答案正确性、验证器评分和步骤级奖励（Shao et al., 2024; Luo et al., 2023a; Chen et al., 2024b; Kazemnejad et al., 2024）。虽然PPO（Schulman et al., 2017）是常用算法，但高计算成本促进了离线强化学习方法的发展（Zhang et al., 2024; Lai et al., 2024; Chen et al., 2024a; Xie et al., 2024; Yuan et al., 2024）。新兴研究直接在基础模型上应用基于准确性的简单奖励（Guo et al., 2025a; OpenAI, 2025），鼓励生成更长、结构化的CoT轨迹。然而，这些方法计算密集，主要适用于大规模模型，在小型模型上的应用仍具挑战性。
Reinforcement learning (RL) has become a widely adopted technique for post-training large language models (LLMs) to better align their outputs with human preferences~\citep{ouyang2022training, achiam2023gpt}. Recent studies extend its application to enhancing reasoning abilities, encouraging longer, structured CoT traces and occasional breakthrough moments~\citep{jaech2024openai, guo2025deepseek}. These approaches treat chain-of-thought (CoT) reasoning as an RL problem, utilizing various reward mechanisms such as final-answer correctness~\citep{xie2025logic,wen2025reinforcement}, verifier-based scoring~\citep{gehringrlef}, and step-level rewards~\citep{zhang2025r1}. While online RL approaches~\citep{schulman2017proximal,shao2024deepseekmath} are commonly used, high computational costs motivate the development of offline RL methods~\citep{zhang2024offlinerl,yuan2024offlinerl}.
% However, these methods remain computationally intensive and are primarily viable for large-scale models, with application to smaller models remaining challenging.
Existing work primarily relies on English-centric data in RL.
% 我们不仅验证了RL在多语言环境下的有效性，更创新性地发现了非英语训练数据在RL框架下的独特优势，并从语言一致性和采样机制两个角度揭示了其背后的工作原理，为多语言模型的训练提供了新的见解。
In this work, we not only validate the effectiveness of RL in a multilingual setting but also innovatively uncover the unique advantages of non-English training data within the RL framework. 
% Furthermore, we reveal the underlying mechanisms from the perspectives of language consistency and sampling processes, offering new insights for training multilingual models.

\section{RL improves the generalization across languages}

% This section presents our comparative results comparing RL and SFT across languages. We detail our experimental setup and then present the main findings that highlight the superiority and the generalization of RL in multilingual reasoning.

% % @shulin Bn 放到后面 对角线对齐
% % @shulin 行最大加粗 列最大斜体

\subsection{Experimental Setup}
\label{sec:exp}
% \paragraph{Base Model and Datasets.} 
\textbf{Base Model and Datasets.} 
We adopt Qwen2.5-3B-Base~\citep{yang2024qwen2} as the base model to clearly explore the impact of RL and SFT. The training datasets are translations of GSM8K~\citep{cobbe2021gsm8k} and LUFFY~\citep{yan2025learning}. We use Qwen3-30B-A3B~\citep{yang2025qwen3} to translate the training data into other languages and utilize the model's verification to further guarantee the quality of the translation. 
To further examine the generality of the observed phenomena, we also include SmolLM3-3B-Base~\citep{bakouch2025smollm3} as an additional base model for verification.
To fully assess the reasoning ability, we evaluate the model on multilingual reasoning benchmarks from three kinds of reasoning: MGSM \citep{shilanguage}, MMath500, and MAIME2024 \citep{son2025linguistic} for mathematical reasoning, MMLU-ProX-Lite \citep{xuan2025mmlu} for commonsense reasoning, and MGPQA-D \citep{qi2025models} for scientific reasoning.
% To fully assess the reasoning ability, we evaluate the model on  multilingual reasoning benchmarks: math reasoning benchmarks including MGSM~\citep{shilanguage}, MMath500 and MAIME2024~\citep{son2025linguistic}, MMLU-ProX-Lite~\citep{xuan2025mmlu} for commonsense reasoning, MGPQA-D for reasoning relative to Scientific questions~\citep{qi2025models}.

% \paragraph{Learning Algorithms and Evaluation Metrics.} 
\textbf{Learning Algorithms and Evaluation Metrics.}
We compare the performance of various tuning algorithms, including SFT and RL. Specifically, we use GRPO to explore the performance of RL. The final answer is explicitly distinguished and encapsulated with in a \verb|\boxed{}|. To evaluate model performance, we calculate the accuracy of each reasoning dataset. We test 6 times for MMath500 and MGSM, and 16 times for MAIME2024. This paper evaluates the reasoning capabilities of LLMs across ten languages: Bengali (Bn), Thai (Th), Swahili (Sw), Japanese (Ja), Chinese (Zh), German (De), French (Fr), Russian (Ru), Spanish (Es), and English (En). 
To measure the relative improvement over the base model's potential, we introduce a generalization score (Gen). This score is calculated by averaging the normalized gains across all test languages, which represents the model's ability to capitalize on the potential for improvement in each language. For a given tuned model $M_{\text{tuned}}$, the generalization score is defined as:
\begin{equation*}
Gen(M_{\text{tuned}}) = \frac{1}{|L|} \sum_{l \in L} \frac{\text{Acc}(M_{\text{tuned}}, l) - \text{Acc}(M_{\text{base}}, l)}{1 - \text{Acc}(M_{\text{base}}, l)}
\end{equation*}
where $L$ is the set of evaluation languages, $\text{Acc}(M, l)$ is the accuracy of model $M$ on language $l$, and $M_{\text{base}}$ is the base model before tuning.
% @shulin 加GEn 公式

% \paragraph{Implementation Details.} 
\textbf{Implementation Details.}
% All experiments employ full-parameter tuning for both SFT and RL to ensure comprehensive capability assessment. Experiments of SFT are conducted using the llamafactory framework, with a learning rate of $1 \times 10^{-5}$ and a batch size of 24. RL methods are implemented using Verl. A consistent configuration is applied across RL methods to ensure an equitable comparison: a learning rate of $1 \times 10^{-6}$, a rollout batch size of 512, a sampling temperature of 1.0, with KL-divergence coefficient is 0.001.
All experiments utilize full-parameter tuning during both the SFT and RL phases to enable a thorough evaluation of model capabilities. The SFT experiments are carried out within the LlamaFactory~\citep{zheng2024llamafactory} framework, employing a learning rate of $2 \times 10^{-5}$, a cosine learning rate scheduler, and a batch size of 32. For RL, the verl~\citep{sheng2024hybridflow} platform is used for implementation. To guarantee a fair comparison among different RL approaches, a uniform set of parameters is adopted: the learning rate is set to $1 \times 10^{-6}$, the rollout batch size to 512, and the sampling temperature to 1.0, along with a KL-divergence coefficient of 0.001.

\subsection{Finding 1: RL exhibits superior cross-lingual generalization than SFT}
% 
% Our empirical analysis demonstrates that reinforcement learning exhibits substantially superior cross-lingual generalization capabilities compared to supervised fine-tuning across all experimental configurations. 

% Our analysis reveals a core finding: RL demonstrates consistently superior cross-lingual generalization capabilities compared to SFT. This advantage is evidenced by (a) {significant performance} improvement in cross-lingual evaluations, (b) {robustness in cross-lingual transfer}, (c) {coherent validation} across datasets, and (d) {effective generalization across languages in} diverse reasoning tasks.

\begin{table*}[t]
 \centering
\caption{%
% Performance of base model, SFT, and RL tuning models on MGSM. Base denotes the orginal Qwen2.5-3B-Base model. SFT (zh) and RL (zh) mean we tune the base model in Chinese data through SFT and RL, respectively. We report the accuracy score on 10 linguistic settings. $\Delta$ (RL-SFT) represents the performance difference between RL and the corresponding SFT score. Each score represents the average accuracy over six measurements. Avg represents the average of the scores of 10 language settings and Gen represents the generalization score. Full table is deferred to  Appendix~\ref{apd:results}.
Performance of base, SFT, and RL models on MGSM.
“Base” denotes Qwen2.5-3B-Base. “SFT (zh)” and “RL (zh)” indicate tuning on Chinese data. We report accuracy on 10 linguistic settings; $\Delta$ (RL–SFT) denotes the performance gap. Each value is averaged over six runs. “Avg” and “Gen” refer to the mean accuracy and generalization score, respectively.
}

 % %%\vspace{-8pt}
\begin{adjustbox}{width=.98\textwidth}{
\begin{tabular}{lcccccccccccc} 
\toprule
Models            & En & Zh     & De & Es & Fr     & Ja & Ru & Th & Sw         & Bn          & Avg         & Gen     \\ 
\midrule
Base & 63.4~           & 48.3~  & 33.5~           & 57.7~           & 38.9~  & 19.5~           & 30.3~           & 17.6~           & 7.3~       & 1.2~        & 31.8~       & 0.0~    \\ 
\midrule
SFT (En)          & 64.7~           & 54.5~  & 50.7~           & 56.4~           & 56.2~  & 36.9~           & 55.5~           & 44.1~           & 6.9~       & 26.2~       & 45.2~       & 18.1~   \\
RL (En)           & 85.8~           & 72.1~  & 70.8~           & 77.3~           & 76.6~  & 61.2~           & 64.9~           & 61.0~           & 9.5~       & 47.5~       & 62.7~       & 49.1~   \\
$\Delta$ (RL-SFT) & +21.1~          & +17.6~ & +20.1~          & +20.9~          & +20.4~ & {+24.3}~ & +9.4~           & +16.9~          & +2.6~      & {{+21.3}}~ & +17.5~      & +30.9~  \\ 
\midrule
SFT (Zh)          & 65.7~           & 58.7~  & 48.4~           & 55.7~           & 56.1~  & 43.5~           & 56.6~           & 45.8~           & 7.5~       & 30.5~       & 46.9~       & 20.4~   \\
RL (Zh)           & 86.1~           & 76.3~  & 74.2~           & 81.1~           & 76.1~  & 64.5~           & 78.1~           & 64.9~           & 10.3~      & 48.3~       & 66.0~       & 52.6~   \\
$\Delta$ (RL-SFT) & +20.4~          & +17.6~ & \textbf{+25.8}~ & \textbf{+25.4}~ & +20.0~ & +21.0~          & +21.5~          & +19.1~          & +2.8~      & +17.8~      & +19.1~      & +32.3~  \\ 
\midrule
SFT (De)          & 63.9~           & 54.2~  & 57.5~           & 55.7~           & 52.8~  & 39.3~           & 55.1~           & 47.6~           & 8.4~       & 28.8~       & 46.3~       & 19.3~   \\
RL (De)           & 91.0~           & 77.6~  & 80.5~           & 82.7~           & 80.0~  & 67.8~           & 81.3~           & 75.3~           & 15.9~      & 63.3~       & 71.5~       & 60.4~   \\
$\Delta$ (RL-SFT) & \textbf{+27.1}~ & \textbf{+23.4}~ & +23.0~          & +27.0~          & \textbf{+27.2}~ & \textbf{+28.5}~          & \textbf{+26.2}~          & \textbf{+27.7}~          & \textbf{+7.5}~ & \textbf{+34.5}~          & \textbf{+25.2}~ & \textbf{+41.2}~  \\ 

\bottomrule
\end{tabular}
}
\end{adjustbox}
 %\vspace{-5pt}
 \label{tab:main table1}
\end{table*}

% As shown in Table~\ref{tab:main table1}, RL consistently outperforms SFT with remarkable margins across all ten languages examined. The performance improvements range from +9.4 points (performance on Russian when trained on English) to +34.5 points (performance on Bengali when trained on German), with an overall average improvement of +17.5 to +25.2 points depending on the training language. 
\textbf{Significant performance improvement.}
% As shown in Table~\ref{tab:main table1}, RL outperforms SFT by remarkable margins across all ten languages examined. The performance improvements range from +9.4 points (evaluating on Russian when trained on English) to +34.5 points (evaluating on Bengali when trained on German), with an overall average improvement of +17.5 to +25.2 points depending on the training language. This establishes a strong baseline for RL's superiority.
As shown in Table~\ref{tab:main table1}, RL consistently outperforms SFT across ten languages. The improvements range from +9.4 points (evaluating Russian when trained in English) to +34.5 points (evaluating Bengali when trained in German), with an overall average improvement of +17.5 to +25.2 points depending on the training language. This establishes a strong baseline for RL's superiority. The complete results are provided in Appendix~\ref{apd:results}.

% Notably, this superiority manifests not only in the training language but, more importantly, in cross-lingual transfer scenarios. For instance, when training on Chinese data, RL achieves 76.3\% accuracy on Chinese evaluation compared to SFT's 58.7\% (+17.6 points), while simultaneously demonstrating superior generalization to other languages such as German (74.2\% vs 48.4\%, +25.8 points) and Spanish (81.1\% vs 55.7\%, +25.4 points). This pattern holds consistently across all training languages, indicating that RL's advantage stems from learning more robust and transferable reasoning representations rather than language-specific optimizations.
\textbf{Robustness in cross-lingual transfer.}
Notably, RL's advantage is most prominent in cross-lingual transfer scenarios, suggesting it learns more robust reasoning strategies rather than optimizing for the training language. For instance, when trained on Chinese, RL not only excels on Chinese evaluation (+17.6 points over SFT) but also generalizes significantly better to typologically distant languages like German (+25.8 points) and Spanish (+25.4 points). The consistency of these improvements across diverse language pairs (e.g., English-Bengali: +21.3 points) indicates that RL fosters the development of multilingual reasoning capabilities.

% The consistency of these improvements across diverse language pairs—including typologically distant languages such as English-Bengali (+21.3 points) and German-Swahili (+7.5 points). This suggests that RL enables models to develop language-agnostic reasoning strategies that generalize effectively across linguistic boundaries.
% 这一结论的有效性不仅限于 mGSM 数据集。如 Table 2 所示，在 MMath500 数据集上，强化学习同样展现了对监督微调的显著优势。例如，在使用中文数据训练时，RL 模型的平均准确率达到了 61.3%，远超 SFT 的 39.8%（高出 +21.5 个百分点）。这种性能上的一致性提升，进一步证实了强化学习能够学习到更具鲁棒性和可迁移性的推理能力，使其在不同的语言和数据集上都具备更强的泛化表现。

\begin{table}[t]
\centering
\caption{Performance of base, SFT, and RL models on MMath500. We report the accuracy score on 6 linguistic settings. 
% Avg represents the average of the scores of 6 language settings and Gen represents the generalization score.
}
 % %%\vspace{-8pt}
\label{tab:main table2}
\begin{adjustbox}{width=.8\textwidth}{
\begin{tabular}{lcccccccc} 
\toprule
Models            & Zh     & Fr     & En     & De     & Ja     & Es     & ~ Avg & Gen  \\ 
\midrule
Base              & 38.9~  & 27.2~  & 49.1~  & 16.3~  & 17.4~  & 36.6~  & 30.9~     & 0.0~            \\ 
\midrule
SFT (En)          & 33.6~  & 56.3~  & 59.7~  & 56.0~  & 18.1~  & 57.2~  & 46.8~     & 22.1~           \\
RL (En)           & 53.7~  & 55.8~  & 62.7~  & 50.9~  & 54.2~  & 56.6~  & 55.7~     & 34.6~           \\
$\Delta$ (RL-SFT) & +20.1~ & -0.4~  & +3.1~  & -5.1~  & +36.1~ & -0.6~  & +8.9~     & +12.5~          \\ 
\midrule
SFT (Zh)          & 41.9~  & 38.4~  & 48.4~  & 35.0~  & 37.2~  & 38.1~  & 39.8~     & 11.3~           \\
RL (Zh)           & 61.3~  & 61.2~  & 63.3~  & 61.2~  & 58.5~  & 62.1~  & 61.3~     & 42.5~           \\
$\Delta$ (RL-SFT) & +19.5~ & +22.8~ & +14.9~ & +26.2~ & +21.3~ & +24.0~ & +21.5~    & +31.2~          \\ 
\midrule
SFT (De)          & 31.7~  & 30.7~  & 38.0~  & 33.5~  & 19.4~  & 30.3~  & 30.6~     & -2.6~           \\
RL (De)           & 61.4~  & 61.5~  & 62.8~  & 60.7~  & 60.1~  & 62.1~  & 61.4~     & 42.6~           \\
$\Delta$ (RL-SFT) & +29.7~ & +30.8~ & +24.8~ & +27.2~ & +40.7~ & +31.8~ & +30.8~    & +45.3~          \\
\bottomrule
\end{tabular}
}
\end{adjustbox}
\vspace{-5pt}
\end{table}

% The validity of this conclusion is not limited to the MGSM dataset. As shown in Table~\ref{tab:main table2}, on the MMath500 dataset, RL demonstrates a similarly significant advantage over SFT. For instance, when trained on Chinese data, the RL model achieves an average accuracy of 61.3\%, far exceeding the 39.8\% of SFT (a +21.5 point improvement). This consistent performance gain further corroborates that RL enables the acquisition of more robust and transferable reasoning abilities, leading to stronger generalization across different languages and datasets.
\textbf{Coherent validation.}
The validity of this conclusion is further strengthened by consistent results on the MMath500 dataset in Table~\ref{tab:main table2}. For example, when trained on Chinese data, the RL model achieves an average accuracy of 61.3\%, substantially surpassing SFT's 39.8\% (a +21.5 point improvement). This cross-dataset corroboration confirms that the enhanced generalization ability of RL is not an artifact of a single benchmark.

\textbf{Effective generalization across languages in other reasoning tasks.}
The superiority of RL extends beyond multilingual mathematical reasoning to challenging out-of-distribution tasks. On benchmarks like MMLU-ProX-Lite and MGPQA-D, RL consistently maintains positive generalization scores, while SFT models often exhibit negative transfer. For instance, on MMLU-ProX-Lite, an RL model trained on German data achieves a generalization score (Gen) of 30.8, starkly contrasting with SFT's 8.0. This demonstrates that the robust reasoning representations acquired via RL are highly transferable across both linguistic and task boundaries. Notably, this advantage holds even under a {double-cross setting}: when {trained on mathematical data in German (De)} and evaluated on a {commonsense reasoning task in Chinese (Zh)}, RL achieves a +20.0 point improvement over SFT. 
% Same phenomenon is observed on SmolLM3-3B-Base, with full results provided in Appendix~\ref{apd:results}.

% This consistent performance improvement across different languages and tasks further confirms that reinforcement learning enables models to acquire more robust and transferable reasoning capabilities, resulting in stronger generalization performance across diverse linguistic and domain settings.
In summary, results from {cross-lingual}, {cross-dataset}, and {cross-task} evaluations robustly support that RL enables models with superior generalization in multilingual reasoning compared to SFT.

\subsection{Finding2: RL using non-English training data yields superior performance to English training data, while SFT does not}

\textbf{Superiority performance gains in non-English RL.} Analyzing the average performance, RL training on non-English data systematically surpasses the English baseline. Specifically, RL trained on German achieves the highest average performance at 71.5\%, followed by French (70.7\%) and Japanese (70.9\%), shown in Table~\ref{tab:main table1 apd} in Appendix~\ref{apd:results}, all substantially exceeding English-based RL training (62.7\%), with the German advantage being a significant +8.8 points.

\begin{table*}[h]
\small
    \centering
    \caption{Performance comparison on {MMLU-ProX-Lite} and {MGPQA-D}. ``Avg'' denotes the average score across languages (En/Zh/De), and ``Gen'' represents the generalization score.}
    \scalebox{0.85}{
    \begin{tabular}{lccccccccccc}
    \toprule 
    \multirow{2}{*}{\textbf{Model}} 
        & \multicolumn{5}{c}{\textbf{MMLU-ProX-Lite}} 
        & \multicolumn{5}{c}{\textbf{MGPQA-D}} \\ 
    \cmidrule(r){2-6} \cmidrule(r){7-11}
        & En & Zh & De & Avg & Gen 
        & En & Zh & De & Avg & Gen \\
    \midrule
    Base             & 9.2  & 2.4  & 3.6  & 5.0  &   0.0   & 21.1 & 20.0 & 20.7 & 20.6 &  0.0     \\
    \midrule
    SFT(En)     & 28.9 & 13.0 & 24.1 & 22.0 & 17.9 & 12.5 & 5.1  & 11.7 & 9.8  & -13.7 \\
    RL(En)      & 40.6 & 31.6 & 25.6 & 32.6 & 29.1 & 30.0 & 23.5 & 25.2 & 26.2 & 7.0   \\
    $\Delta$ (RL-SFT)  & +11.6 & +18.6 & +1.6  & +10.6 & +11.2 & +17.4 & \textbf{+18.4} & +13.5 & +16.4 & +20.7  \\
    \midrule
    SFT(Zh)     & 24.4 & 11.6 & 21.3 & 20.3 & 7.4 & 20.7 & 18.1 & 14.2 & 17.7 & -3.7  \\
    RL(Zh)      & 40.8 & 35.0 & 34.4 & 36.7 & 33.4 & 25.0 & 27.3 & 28.3 & 26.9 & 7.8   \\
    $\Delta$ (RL-SFT)  & +16.3 & \textbf{+23.4} & +13.1 & +16.4 & +19.8 & +4.3  & +9.2  & +14.1 & +9.2  & +11.5  \\
    \midrule
    SFT(De)     & 15.8 & 7.3  & 15.0 & 12.7 & 8.0  & 7.2  & 12.8 & 8.8  & 9.6  & -13.9 \\
    RL(De)      & 39.9 & 27.2 & 35.5 & 34.2 & 30.8 & 26.2 & 27.2 & 25.3 & 26.2 & 7.1   \\
    $\Delta$ (RL-SFT)   & \textbf{+24.1} & +20.0 & \textbf{+20.5} & \textbf{+21.5} & \textbf{+22.8} & \textbf{+19.0} & +14.4 & \textbf{+16.6} & \textbf{+16.7} & \textbf{+21.0}  \\
    \bottomrule
    \end{tabular}
    }
    \label{tab:mmlu-gpqa}
    % \vspace{-10pt}
\end{table*}
% 
% This superiority is particularly evident in cross-lingual transfer scenarios. For example, RL training on German not only achieves exceptional performance on German evaluation (80.5\%) but also demonstrates remarkable transfer to distant languages such as Bengali (63.3\% vs 47.5\% for English training, +15.8 points) and Thai (75.3\% vs 61.0\% for English training, +14.3 points). Similar patterns emerge across other non-English training languages, with each consistently outperforming English-based training in both within-language and cross-lingual evaluation scenarios.
% \textbf{Enhanced cross-lingual transfer through non-English RL.} 
Further more, the superiority is also pronounced in cross-lingual scenarios. For example, RL trained on German not only excels on German evaluation (80.5\%) but also shows remarkable transfer to distant languages like Bengali (63.3\% vs 47.5\% for English, +15.8 pts) and Thai (75.3\% vs 61.0\% for English, +14.3 pts), indicating learning of transferable representations.

% Crucially, this phenomenon is specific to RL paradigms. Examination of the SFT results reveals no such systematic advantage for non-English training languages, with performance variations remaining within statistical noise levels (46.3-47.6 average performance). 
% \textbf{A Clear Dichotomy with SFT.} This phenomenon is exclusive to RL. In stark contrast, SFT results show minimal variations, from Avg scores as 46.3\% on German to 47.6\% on Japanese, across different training languages, shown on Table~\ref{tab:main table1 apd} in Appendix~\ref{apd:results}, with performance differences remaining within statistical noise, which rules out data quality as a sole explanation and underscores the critical role of the RL objective.

% 非英语训练语言在强化学习中的优势同样在 MMath500 数据集（Table 2）中得到了验证。具体来说，使用德语数据训练的 RL 模型取得了 71.5% 的最高平均性能，显著优于在中文（66.0%）和英文（62.7%）上训练的模型。
% As shown in Table~\ref{tab:main table2}, the superiority of non-English training languages in RL is also validated on MMath500 dataset. Specifically, the RL model trained on German data achieves the highest average performance at 71.5\%, significantly outperforming models trained on Chinese (66.0\%) and English (62.7\%). This result stands in contrast to the performance of SFT, which shows an
% opposite trend (46.8\%, 39.8\%, and 30.6\% on English, Chinese, and German training data).
\textbf{Similar phenomenon across diverse reasoning tasks.} The pattern is consistently validated on diverse benchmarks. On MMLU-Pro-X Lite, from Table~\ref{tab:mmlu-gpqa}, RL trained on Chinese achieves 36.7\%, outperforming RL trained on English (32.6\%). On MGSM, RL trained on German attains a generalization score of 41.2\%, significantly higher than RL trained on English (30.9\%), confirming the robust and generalizable benefits of non-English RL training.

\textbf{The Different Phenomenon observed in SFT.}
This phenomenon appears exclusively with RL. In contrast, SFT results exhibit minimal variation across training languages, with Avg scores ranging from 46.3\% on German to 47.6\% on Japanese (see Table~\ref{tab:main table1 apd} in Appendix~\ref{apd:results}). The performance differences remain within statistical noise, ruling out data quality as the sole explanation and highlighting the critical role of the RL objective. 

\subsection{Same phenomenon on another base model}
In Table~\ref{tab:smol table}, we report the performance of {SmolLM3-3B-Base} under the same configuration of {Qwen2.5-3B-Base}. We find that our observations are consistently.

\textbf{Finding 1: RL exhibits superior cross-lingual generalization than SFT.}  
Across all training languages, RL consistently and significantly outperforms SFT. The improvements are substantial, ranging from +11.7 points (evaluating Swedish when trained on German) to +34.7 points (evaluating Chinese when trained on German). 
% The average gains remain high across all settings, from +26.1 (RL (En)) to +28.1 (RL (De)). For instance, RL (De) achieves +27.7 on German, +34.0 on Russian, and +31.9 on French, demonstrating robust cross-lingual transfer.

\textbf{Finding 2: RL using non-English training data yields superior performance to English training data, while SFT does not.}  
Similar to the trend observed on {Qwen2.5-3B-Base}, RL trained on non-English data surpasses RL trained on English. RL (De) reaches the highest average accuracy at 69.9 and the strongest generalization score at 64.9.
% , followed by RL (Zh) with 68.7 (Avg) and 63.4 (Gen), both clearly outperforming RL (En) (64.6 / 58.6). 
In contrast, SFT models remain far behind.
% (e.g., SFT (En) Avg: 38.6).
% and their non-English variants only show marginal gains, underscoring that the non-English advantage is uniquely enabled by RL optimization rather than data alone.

\begin{table*}[t]
 \centering
\caption{Performance of base model, SFT, and RL tuning models on MGSM. Base denotes the orginal SmolLM3-3B-Base model. We report the accuracy score on 10 linguistic settings. }
 \begin{adjustbox}{width=.98\textwidth}{
\begin{tabular}{lcccccccccccc} 
\toprule
Models            & En & Zh     & De & Es & Fr     & Ja & Ru & Th & Sw         & Bn          & Avg         & Gen     \\ 
\midrule
Base    & 29.3          & 17.1          & 24.3          & 27.3          & 26.9          & 18.1          & 24.1          & 13.0          & 4.0           & 5.7           & 19.0           & 0.0           \\
\midrule
SFT (En)& 60.3          & 40.7          & 49.3          & 51.8          & 48.0          & 32.8          & 47.7          & 37.7          & 7.3           & 10.0          & 38.6          & 25.3          \\
RL (En) & 87.3          & 70.8          & 78.3          & 81.0          & 81.3          & 62.2          & 79.7          & 67.3          & 16.5          & 22.2          & 64.6          & 58.6          \\
$\Delta$ (RL-SFT)       & +27.1          & +30.1          & \textbf{+28.9} & +29.2          & \textbf{+33.3} & +29.4          & +32.0          & +29.5          & +9.1           & +12.2          & +26.1          & +33.3          \\
\midrule
SFT (Zh)& 58.9          & 51.5          & 50.5          & 54.9          & 54.9          & 39.3          & 48.7          & 43.7          & 9.2           & 11.7          & 42.3          & 30.0          \\
RL (Zh) & 88.6          & 77.1          & 79.3          & 82.5          & 80.9          & 71.1          & 82.4          & 76.2          & 20.1          & 28.5          & 68.7          & 63.4          \\
$\Delta$ (RL-SFT)      & \textbf{+29.7} & +25.6          & +28.7          & +27.6          & +26.0          & \textbf{+31.8} & +33.7          & +32.5          & +10.9          & +16.9          & +26.3          & +33.4          \\
\midrule
SFT (De)& 60.1          & 43.5          & 53.9          & 56.1          & 52.1          & 38.3          & 51.8          & 43.3          & 9.0           & 9.7           & 41.8          & 29.4          \\
RL (De) & 85.1          & 78.2          & 81.7          & 85.6          & 84.1          & 69.1          & 85.8          & 77.2          & 20.7          & 31.3          & 69.9          & 64.9          \\
$\Delta$ (RL-SFT)      & +24.9          & \textbf{+34.7} & +27.7          & \textbf{+29.5} & +31.9          & +30.8          & \textbf{+34.0} & \textbf{+33.9} & \textbf{+11.7} & \textbf{+21.7} & \textbf{+28.1} & \textbf{+35.5} \\

\bottomrule
\end{tabular}
}
\end{adjustbox}
\label{tab:smol table}
\vspace{-5pt}
\end{table*}

\section{Mechanics Analysis of RL's Cross-lingual Generalization}
% \vspace{-5pt}
% 为了探究为什么RL比SFT拥有更好的泛化性，以及为什么在非英语语言上的RL训练会带来比英语语言上的RL训练更好的性能，我们查看RL训练后的模型在测试数据上的回答，如图1，我们发现，使用德语问题以及德语instruction来进行RL训练，训练好的模型在回答的阶段并不会完全遵照德语来回答，反而会用非德语或者混杂的语言来进行思考。这引起了我们的好奇，并给出一个假设，是否是因为这种不一致的语言思考，导致RL的训练过程具备更好的泛化性呢？
% 为了尝试验证这个假设，我们分别使用prompt强限制语言的方式和对RL训练增加语言一致性的reward来分别做对比实验。

% To investigate why RL exhibits superior generalization capabilities compared to SFT, and why RL training on non-English languages yields better performance than RL training on English, we examine the part of responses (full responses in Appendix~\ref{apd:case}) generated by RL-trained models on test data, due to space limits, as illustrated in Table~\ref{tab:samples}. 
To investigate why RL exhibits stronger generalization than SFT, and why RL training on non-English data outperforms that on English, we present an abbreviated set of responses generated by RL-trained models on the test set in Table~\ref{tab:samples}. The complete responses are provided in Appendix~\ref{apd:case}.
Our analysis reveals that when models are trained using German instructions during RL training, the resulting models do not strictly adhere to German when generating thinking and responses. Instead, they employ non-German or mixed languages for reasoning processes. This observation attracts our attention and leads us to propose a hypothesis: could this inconsistent language usage in reasoning contribute to the enhanced generalization observed in RL training?

\subsection{Exploration of language consistency in RL}
% To empirically validate this hypothesis, we conduct comparative experiments using two distinct approaches: (1) employing prompts that strictly constrain language usage, and (2) incorporating language consistency rewards—which encourage the model to adhere to the language of the question—into the RL training process. 
% incorporating language consistency rewards (adhere to the language of the question) into the RL training process as follows:
To empirically validate this hypothesis, we conduct comparative experiments using two distinct approaches: (1) employing prompts that strictly constrain language usage, and (2) incorporating the language consistency reward that encourages the model to adhere to the language of the question into the RL training process. The details are as follows:
 % \vspace{-10pt}
\begin{equation}
    r_{\text{overall}} = 0.5 r_{\text{acc}} + 0.5 r_{\text{consistency}} \label{eq:reward}
\end{equation}
 
% \begin{CJK*}{UTF8}{gbsn}
\begin{table*} [t]
\centering
\caption{Case study of on MMath500. We present a case of German question.}
 %%\vspace{-8pt}
\label{tab:samples}
\begin{adjustbox}{width=.75\textwidth}{
\begin{tabular}{l|l} 
\toprule
    \multicolumn{2}{l}{\begin{tabular}[c]
    {@{}p{0.96\linewidth}@{}}
    \sethlcolor{blue!10}\hl{\textbf{Question:}} Ein Matheclub veranstaltet einen Kuchenverkauf als Spendenaktion, um Geld für eine bevorstehende Reise zu sammeln. Sie verkaufen $54$ Kekse zu drei für \$1\$, und \$20\$ Cupcakes zu \$2\$ pro Stück, und \$35\$ Brownies zu \$1\$ pro Stück. Wenn es den Matheclub \$15\$ gekostet hat, diese Artikel zu backen, wie hoch war ihr Gewinn? \\
    \textbf{Answer}: 78\\
    \end{tabular}} \\ 
\midrule
    \multicolumn{2}{l}{\begin{tabular}[c]{@{}p{0.96\linewidth}@{}}
    \sethlcolor{blue!10}\hl{\textbf{Base:}}\\
    % \textbf{Original Awareness Information}: 
    Da Kekse zu drei für $\$1$ verkaufen werden, müssen wir zahlenweise Kekse zählen. Es gibt $54$ Kekse zu backen, und jeder Kuchen besteht aus drei Keksen. Also haben sie $54/3=18$ Kuchen zu backen. 
    $\ \ldots\ $
    % Zudem haben sie $20$ Cupcakes zu backen und $35$ Brownies zu backen. Die Matheclub hat also insgesamt $18+20+35=73$ Artikel zu backen. Sind sie zu backen, hat es ihnen $73 \times 2 = \$146$ gekostet.
    Da sie für die Backzeit $15$ Dollar ausgeben, haben sie für den Verkauf $146-15 = \$131$ eingenommen. Daher haben sie einen Gewinn von $\boxed{131}$.
    \end{tabular}} \\ 
\midrule
    \multicolumn{2}{l}{\begin{tabular}[c]{@{}p{0.96\linewidth}@{}}
    \sethlcolor{blue!10}\hl{\textbf{Reinforcement Learning (with German Data):}}\\
    % \textbf{Original Awareness Information}: 
    
    1. Calculate the revenue from Kekse:
       - They sold 54 Kekse at a rate of 3 Kekse for \$1.
       - The number of sets of 3 Kekse in 54 is \( \frac{54}{3} = 18 \).
       - Revenue from Kekse = \( 18 \times 1 = 18 \) dollars.
    
    $\quad \ldots\ $
    % 2. Calculate the revenue from Cupcakes:
    %    - They sold 20 Cupcakes at \$2 each.
    %    - Revenue from Cupcakes = \( 20 \times 2 = 40 \) dollars.
    
    % 3. Calculate the revenue from Brownies:
    %    - They sold 35 Brownies at \$1 each.
    %    - Revenue from Brownies = \( 35 \times 1 = 35 \) dollars.
    
    % 4. Calculate the total revenue:
    %    - Total revenue = Revenue from Kekse + Revenue from Cupcakes + Revenue from Brownies
    %    - Total revenue = \( 18 + 40 + 35 = 93 \) dollars.
    
    5. Calculate the profit:
       - Cost to make the items = \$15.
       - Profit = Total revenue - Cost to make the items
       - Profit = \( 93 - 15 = 78 \) dollars.
    
    Therefore, the profit is \(\boxed{78}\).
    \end{tabular}} \\ 
\midrule
    \multicolumn{2}{l}{\begin{tabular}[c]{@{}p{0.96\linewidth}@{}}
    \sethlcolor{blue!10}\hl{\textbf{Reinforcement Learning (with German Data and with Consistency Reward):}}\\
    Zuerst berechnen wir den Einnahmen von den Keksen. Da sie drei Kekse für $\$1$ verkaufen, erhalten sie:
    $54 \div 3 = 18 \text{ Doppelsätze}$,
    $\quad\ldots\ $
    % $18 \times 2 = 36 \text{ Dollar}$
    % Für die Cupcakes erhalten sie:
    % $20 \div 2 = 10 \text{ Stück}$, $10 \times 2 = 20 \text{ Dollar}$
    % Für die Brownies erhalten sie:
    % $35 \div 1 = 35 \text{ Stück}$, $35 \times 1 = 35 \text{ Dollar}$
    % Die Gesamtsumme ihrer Einnahmen beträgt:
    % $36 + 20 + 35 = 91 \text{ Dollar}$
    Da es ihnen $\$15$ gekostet hat, um die Artikel zu backen, erhalten sie:
    $91 - 15 = 76 \text{ Dollar}$
    Die Gewinnsumme beträgt \boxed{76}.
    \end{tabular}} \\
% \midrule
\bottomrule
\end{tabular}
}\end{adjustbox}
% \vspace{-10pt}
\end{table*}
% \end{CJK*}

% \vspace{-27pt}
% \vspace{-5pt}

% \vspace{-20pt}
\begin{figure}[t!]
    % \vspace{-5pt}
    % ------------------- 左边的图形 -------------------
    \begin{minipage}[c]{0.50\textwidth}
        \centering
        % 假设图片文件名是 luffy_mmath500.pdf
        \includegraphics[width=\linewidth]{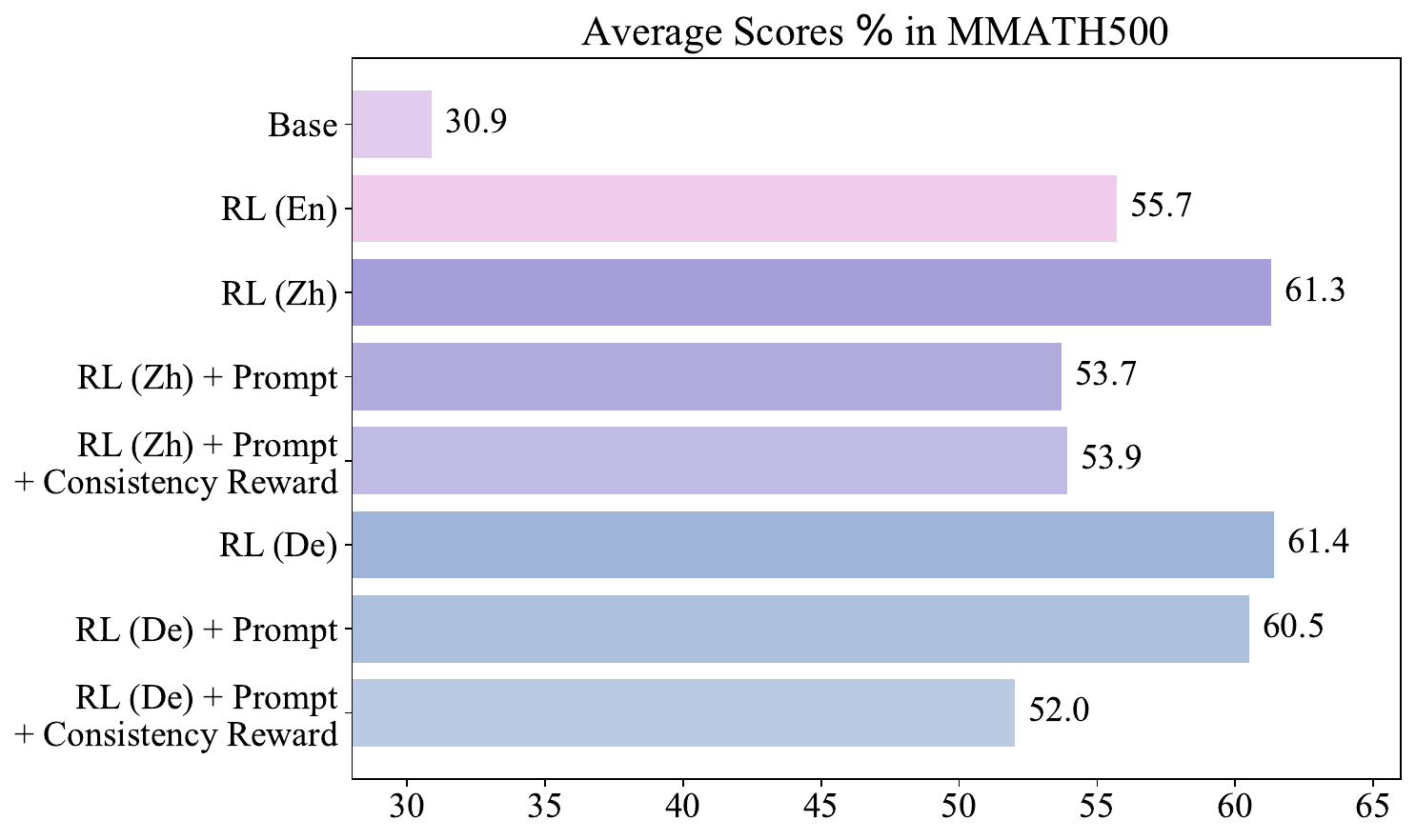} 
        % 调整后的图标题
        \captionof{figure}{Scores on MMath500. The chart compares the average accuracy of different models. “RL (Zh)” indicates training on Chinese data.}
        \label{fig:mmath500} % 使用新的label
    \end{minipage}
    \hfill % 在左右两个盒子之间 ایجاد一个弹性空白
    % ------------------- 右边的表格 -------------------
    \begin{minipage}[c]{0.47\textwidth}
        \centering
        \captionof{table}{Language consistency of models on MMath500. We test 6 times and report the average percentage of language consistency.}
        % 使用 adjustbox 来精确控制表格宽度
        \begin{adjustbox}{width=\linewidth}
            \begin{tabular}{lccc} 
            \toprule
            Models                          & En (\%) & Zh (\%) & De (\%)  \\ 
            \midrule
            Base                            & 99.4    & 91.4    & 94.5     \\
            SFT (En)                        & 98.6    & 99.3    & 83.7     \\
            RL (En)                         & 99.9    & 89.0    & 96.2     \\ 
            \midrule
            SFT (Zh)                        & 99.7    & 98.9    & 81.3     \\
            RL (Zh)                         & 99.8    & 0.0     & 0.0      \\
            + Prompt                        & 99.3    & 99.6    & 97.3     \\
            + Prompt + Consistency Reward & 99.9    & 99.8    & 98.9     \\ 
            \midrule
            SFT (De)                        & 94.2    & 85.5    & 99.1     \\
            RL (De)                         & 99.7    & 4.8     & 0.0      \\
            + Prompt                        & 99.8    & 52.4    & 0.0      \\
            + Prompt + Consistency Reward & 99.8    & 99.8    & 99.9     \\
            \bottomrule
            \end{tabular}
        \end{adjustbox}
        % 调整后的表标题
        %%\vspace{7pt}
        
        \label{tab:language consistency} % 使用新的label
    \end{minipage}
    \vspace{-10pt}
\end{figure}

% \v(-15pt)
% 排除数据对结果的影响 做Rej-FT

% 为了验证前述假设——即强化学习训练中不一致的语言使用有助于提升模型的泛化能力——我们进行了深入的实验分析。我们通过强制模型在推理时使用特定语言，并观察其性能变化，来探究语言一致性的具体影响。实验结果清晰地展示在表3和表4中，表5则提供了一个具体的案例。

% 强制语言一致性导致性能下降。 从表3的平均准确率（Average Accuracy）来看，无论是使用中文数据训练的RL (Zh)模型（61.3%），还是使用德文数据训练的RL (De)模型（61.4%），其原始性能都非常出色。然而，当我们通过添加提示词（"+ Prompt"）来强制模型使用与训练数据一致的语言进行推理时，模型的平均性能出现了显著的下滑（RL (Zh)降至53.7%，RL (De)降至60.5%）。更有甚者，当我们进一步引入语言一致性奖励（"+ Prompt + Consistency Reward"）来更严格地约束语言使用时，性能下降得更为明显，尤其是在RL (De)模型上，平均准确率骤降至52.0%。这一结果有力地证明，强制的语言一致性反而损害了模型的跨语言推理能力。

% 语言不一致是高性能模型的普遍特征。 表4量化了各个模型输出的语言一致性比例。数据显示，未经约束的RL (Zh)和RL (De)模型在各自训练语言（中文和德文）上的一致性极低（均为0.0%），这表明它们在解决问题时，几乎完全依赖其他语言（主要是英语）进行思考和推理。与之形成鲜明对比的是，添加了约束条件的模型，其语言一致性显著提高（例如，RL (De)在添加一致性奖励后，德语一致性达到99.9%）。结合表3的性能数据，我们可以得出一个关键结论：正是这种语言上的“不一致性”，即模型自发地采用一种更具普适性的“中间语言”（lingua franca）进行推理，才是其获得强大跨语言泛化能力的根源。

% 案例分析（表5）直观展示了这一现象。 在处理一个德语问题时，未经约束的RL (De)模型完全使用英语逐步推导，并得出了正确答案。然而，被施加了一致性奖励的模型则全程使用德语进行推理，但其逻辑步骤出现偏差，最终导致了错误的答案。这个案例生动地揭示了，强制模型使用特定语言进行复杂的逻辑推理，可能会限制其调用在预训练阶段学到的、更强大的、与语言无关的推理模块。

% 综上所述，强化学习之所以能获得卓越的跨语言泛化能力，一个关键机制在于它允许模型在推理时“摆脱”特定训练语言的束缚，转而使用一种潜在的、更通用的语言（如英语）来构建思维链。这种语言上的灵活性和不一致性，恰恰是其成功的关键。

% To validate the hypothesis that inconsistent language use during RL training enhances generalization, we conduct an in-depth experimental analysis. 
% \vspace{-15pt}
We investigate the impact of language consistency by forcing the model to use a specific language during inference and observing performance changes. The results are presented in Figure~\ref{fig:mmath500} and Table~\ref{tab:language consistency}, with a specific case study provided in Table~\ref{tab:samples}.

We observe two key aspects:
(a) Language inconsistency serves as a potential source of cross-lingual generalization capability, and (b) Building upon this mechanism, RL (De) exhibits greater language inconsistency than RL (En), resulting in superior cross-lingual performance. These observations suggest that the degree of flexibility in deviating from training language constraints may determine the extent of cross-lingual generalization achieved by RL-trained models.

% Enforcing language consistency leads to performance degradation. As shown by the average accuracy in Figure~\ref{fig:mmath500}, the baseline performance of both the RL (Zh) model trained on Chinese data (61.3\%) and the RL (De) model trained on German data (61.4\%) is exceptionally strong. However, when we add a prompt (``+ Prompt'') to compel the models to use the same language as the training data for reasoning, their average performance significantly decline (to 53.7\% for RL (Zh) and 60.5\% for RL (De)). Furthermore, when we introduce a language consistency reward (``+ Prompt + Consistency Reward'') for stricter enforcement, the performance drops even more dramatically, particularly for the RL (De) model, whose average accuracy plummets to 52.0\%. This outcome strongly demonstrates that enforced language consistency impairs the model's cross-lingual reasoning capabilities.
%
%
\begin{figure*}[t]
 %%\vspace{-10pt}
    \centering
    \small
    
    % \subfloat[A comparison of performance on MGSM.]{
    \includegraphics[width=0.70\textwidth]{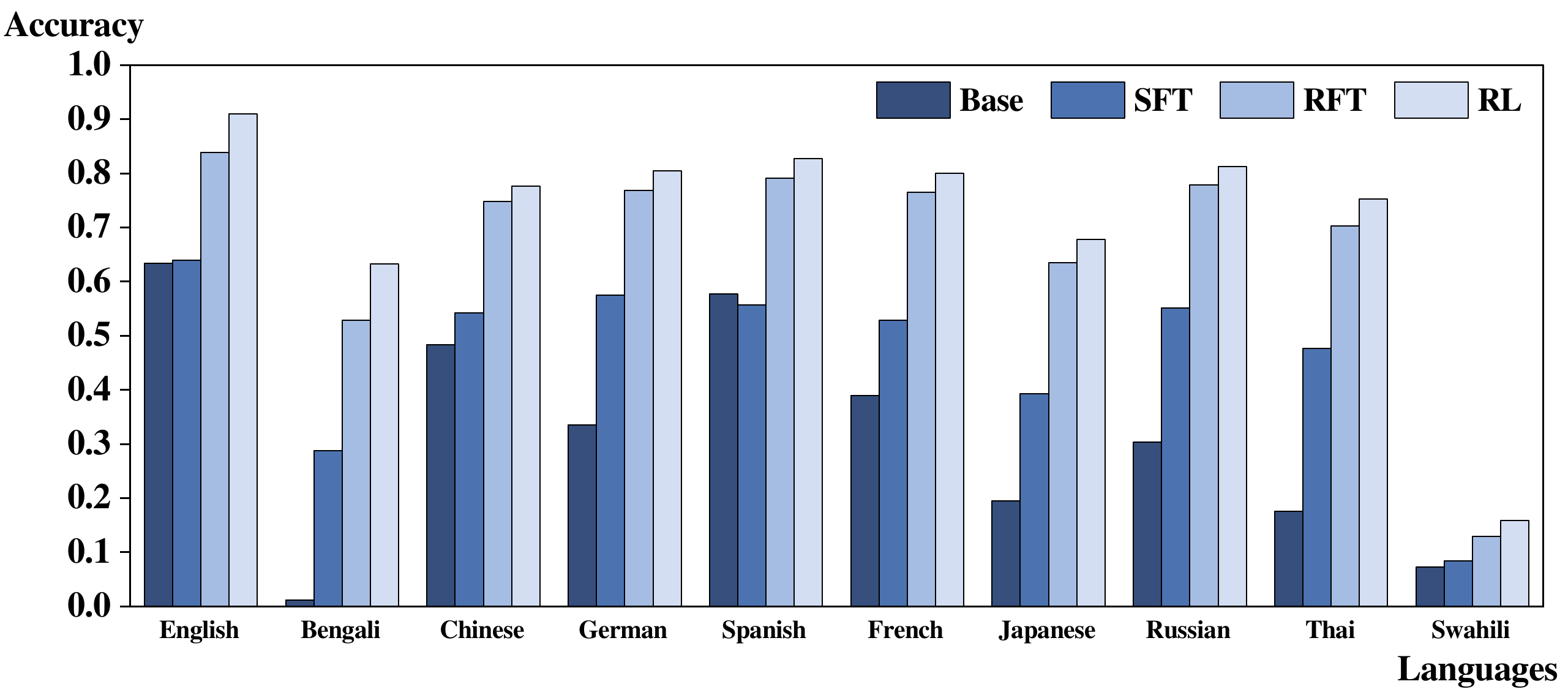}
    \label{fig:bar1}
    % }
    \caption{Model performance comparisons among the Base, SFT, RFT, and RL models on MGSM. We use German data in LUFFY in SFT, RL, RFT for training.}
    \label{fig:combined}
     \vspace{-10pt}
\end{figure*}
%
%
% Language inconsistency is a common trait of high-performing models. We utilize langid~\citep{yang2025navajo,lui2012langid} to calculate the consistency score. Table~\ref{tab:language consistency} quantifies the language consistency ratio of the models' outputs. The data reveals that the unconstrained RL (Zh) and RL (De) models exhibit extremely low consistency in their respective training languages (both 0.0\%). 
% This indicates that when solving problems, they almost entirely rely on other languages for their thought processes. In contrast, models with constraints show a significant increase in language consistency (e.g., RL (De) with the consistency reward reaches 99.9\% consistency in German). 
\textbf{Language inconsistency enhances cross-lingual generalization.}
Enforcing language consistency significantly degrades performance. As demonstrated in Figure~\ref{fig:mmath500}, both RL (Zh) and RL (De) models achieve strong baseline performance (61.3\% and 61.4\%, respectively). However, when constrained to use their training languages through prompting, performance drops substantially—RL (Zh) falls to 53.7\% and RL (De) to 60.5\%. The degradation becomes even more pronounced when language consistency rewards are applied, with RL (De) plummeting to 52.0\%. This pattern strongly indicates that enforced language consistency impairs cross-lingual reasoning capabilities.

% This indicates that when solving problems, they tend to rely on other languages for their thought processes. In contrast, models with constraints show a significant increase in language consistency. Specifically, the model that is trained on German data with the consistency reward reaches a very high consistency rate of 99.9\%.
% By correlating this with the performance data in Table~\ref{tab:control language}, we can draw a key conclusion: this linguistic inconsistency—the model's spontaneous adoption of a more universal ``lingua franca'' for reasoning—is the root of its powerful cross-lingual generalization ability.
% By correlating this with the performance data in Table~\ref{tab:main table2}, we can draw a key conclusion: the model's powerful cross-lingual generalization ability stems from a linguistic inconsistency—its spontaneous adoption of a universal ``lingua franca'' for reasoning.

Table~\ref{tab:language consistency} reveals that unconstrained RL models show low consistency in their training languages—both RL (Zh) and RL (De) achieve 0.0\% consistency, indicating they spontaneously adopt other languages during reasoning. Conversely, constrained models exhibit high consistency rates (up to 99.9\% for RL (De) with consistency rewards), but at the cost of reduced performance. This negative correlation between language consistency and performance suggests that linguistic flexibility enables models to leverage more powerful, multilingual reasoning modules.

% The case study in Table~\ref{tab:samples} offers a direct illustration. When tasked with a German problem, the unconstrained RL (De) model reasons in English-German mixtured language and arrived at the correct answer.
% The case study in Table~\ref{tab:samples} offers a direct illustration: when tasked with a German problem, the unconstrained RL (De) model reasons in a mixture of English and German and arrives at the correct answer.
% In contrast, the model subjects to the consistency reward reason only in German, but its logical steps are flawed, leading to an incorrect answer. This case vividly illustrates that forcing a model to use a specific language for complex logical reasoning may inhibit its ability to leverage the more powerful, language-agnostic reasoning modules acquired during pre-training.
\textbf{Case analysis of the language inconsistency.} As shown in Table~\ref{tab:samples}, When solving German question, the unconstrained model (RL (De)) employs mixed English and German reasoning and reaches correct solutions, while the consistency-constrained model, despite only reasoning in German, produces flawed logical steps and wrong answers. This demonstrates that constraining models to specific languages may inhibit access to more robust reasoning patterns established during pre-training.

% 相较之下，监督微调（SFT）模型则表现出截然不同的行为。从Table 4可以看出，SFT模型（如SFT (De)）在处理德语问题时，其输出的语言一致性高达99.1%。这种“指令遵循”的行为源于SFT的训练机制——严格模仿高质量的、语言统一的训练样本。然而，这种对特定语言模式的忠实模仿恰恰限制了其泛化能力。当面对其他语言的问题时，模型难以摆脱在训练中形成的语言-思维定式，从而导致跨语言性能受损。因此，RL通过允许甚至鼓励语言不一致的“自由思考”获得了更强的泛化能力，而SFT则因其模仿范式受困于语言的牢笼。

% In contrast, SFT models exhibit different behavior. As shown in Table~\ref{tab:language consistency}, an SFT model like SFT (zh) maintains a language consistency of 99.7\% when processing Chinese questions. This ``instruction-following'' behavior stems from the SFT mechanism, which strictly imitates high-quality, linguistically uniform training samples. However, this faithful imitation of specific linguistic patterns is precisely what constrains its generalization ability. When faced with problems in other languages, the model struggles to break free from the language-thought paradigm established during training, leading to impaired cross-lingual performance. Therefore, RL achieves stronger generalization by permitting, and even encouraging, linguistically inconsistent free thinking, whereas SFT remains trapped in a linguistic cage by its imitation-based paradigm.
\textbf{Language inconsistency in non-English RL.}
The performance comparison in Figure 2 shows that RL (De) achieves 61.4\% average accuracy compared to RL (En)'s 55.7\%. More importantly, RL (De) maintains strong performance across different target languages, while RL (En) shows more pronounced degradation in non-English tasks. This suggests that German-based training may provide advantages for cross-lingual generalization.

Different source languages yield distinct generalization patterns. The superior performance of RL (De) may stem from German's grammatical complexity and its linguistic distance from other languages, potentially encouraging the development of more multilingual reasoning strategies. In contrast, English-based training might lead to more language-specific reasoning patterns that transfer less effectively across languages.

\textbf{Language consistency in SFT.} Unlike RL models, SFT models exhibit high language consistency (e.g., SFT (Zh) maintains 99.7\% consistency) due to their imitation-based training paradigm. While this consistency appears desirable, it actually constrains generalization by trapping models within language-specific thought patterns established during training, leading to impaired cross-lingual performance when facing problems in other languages.

% In summary, a key mechanism behind RL's superior cross-lingual generalization is its ability to break free from the constraints of the specific training language during reasoning, instead utilizing a latent, inconsistent language to construct its chain of thought. This linguistic flexibility and inconsistency is a critical component of its success.
% Our findings reveal that RL's superior cross-lingual generalization stems from its ability to break free from training language constraints during reasoning, utilizing inconsistent but flexible language patterns to construct reasoning chains. This linguistic inconsistency, rather than being a limitation, represents a critical mechanism that enables models to access more powerful, multilingual reasoning capabilities. Furthermore, the choice of training language matters—German-based RL training demonstrates superior generalization compared to English-based approaches, highlighting the importance of considering linguistic diversity in RL training design.

\subsection{Exploration of sampling in RL}
% SFT RFT RL 公式比较 @yiran

% 为了进一步探究强化学习（RL）相比监督微调（SFT）的优势来源，我们分析了采样（Sampling）在提升模型性能中的作用。在图2中，我们引入了拒绝采样微调（Rejection Sampling Fine-Tuning, RFT）作为SFT和完整RL之间的中间参照。RFT通过从模型中多次采样，并仅保留能够得出正确答案的样本进行微调，代表了一种比SFT更进一步的探索机制。

% 图2中的三个子图（MGSM, MMath500, MAIME-2024）一致地揭示了一个清晰的性能递进关系：Base < SFT < RFT < RL。这一趋势有力地说明了模型对解题路径的探索是提升其推理能力的核心驱动力。

% 从SFT到RFT的性能飞跃证明了探索的价值。 SFT仅仅是模仿数据集中给定的单一正确解题路径，而RFT则通过采样，让模型有机会接触到更多样的、可能成功或失败的推理链条。通过筛选出成功的路径并进行学习，模型能够更好地理解何为“正确的推理模式”，而不仅仅是记忆一个固定的解法。图2中，RFT在所有数据集和多数语言上都显著优于SFT，这表明，仅仅是引入“探索-筛选”这一简单机制，就能有效增强模型的泛化推理能力。

% 从RFT到RL的持续提升则凸显了策略优化的重要性。 RFT是一种被动的筛选，而RL（如此处使用的GRPO算法）则是一种主动的策略优化。RL不仅要找到正确的答案，还要通过策略梯度等方式，主动调整模型参数，使其更有可能在未来生成高奖励（即正确）的推理路径。图2显示，RL的性能始终是四种方法中最高的，这说明通过强化学习对生成策略进行端到端的优化，能比简单的拒绝采样更有效地提升模型内在的、可泛化的推理逻辑。

% 因此，我们可以得出结论，强化学习的优越性不仅来自于其奖励机制，更深层次地源于其内在的“探索-优化”循环。通过广泛采样来探索问题的解空间，并利用奖励信号来主动优化生成高价值推理路径的策略，模型得以学习到更加鲁棒和通用的问题解决方法，这种方法超越了单一语言或特定样本的限制，从而实现了卓越的跨语言泛化。

To further investigate the source of RL's advantage over SFT, we analyze the role of sampling in performance enhancement. We introduce Rejection Sampling Fine-Tuning (RFT)~\citep{touvron2023llama} as an intermediate baseline between SFT and full RL. 
The RFT we use involves sampling multiple times from the model after RL training.
It then fine-tunes the model using only the samples that yield the correct answer. This represents a more on-policy exploration mechanism than SFT.

As shown in Figure~\ref{fig:combined}, across all languages, accuracy increases progressively from the Base model to SFT, RFT, and finally to the RL-tuned model. Specifically, SFT achieves 46.3\% average accuracy, RFT improves to 66.8\%, and RL reaches 71.5\%. This trend underscores the importance of the model's exploration of solution paths in enhancing its reasoning abilities.

% \paragraph{Better Performance with Data Aligned to the Model's Distribution.}
\textbf{Better Performance with Data Aligned to the Model's Distribution.}
Although SFT follows a completely correct off-policy solution path, RFT data, more aligned with the model's distribution, enables the model to explore reasoning chains better suited to its own configuration through sampling. This alignment helps the model capture reasoning patterns and optimization trajectories more effectively, allowing it to generalize beyond memorized solutions.

% \paragraph{The Importance of Online Optimization in RL.}
\textbf{The Importance of Online Optimization in RL.}
Compared to RFT, RL (GRPO in our experiments) continuously performs more on-policy sampling with both positive and negative examples during training. This not only further aligns the data with the model but also goes beyond mere imitation learning. As shown in Figure~\ref{fig:combined-apd}, RL consistently outperforms the other methods across all languages, demonstrating that the online policy optimization process in RL is more effective at enhancing the model's generalizable reasoning than RFT.

\begin{figure}[t]
 %%\vspace{-10pt}
    \centering
    \includegraphics[width=0.8\textwidth]{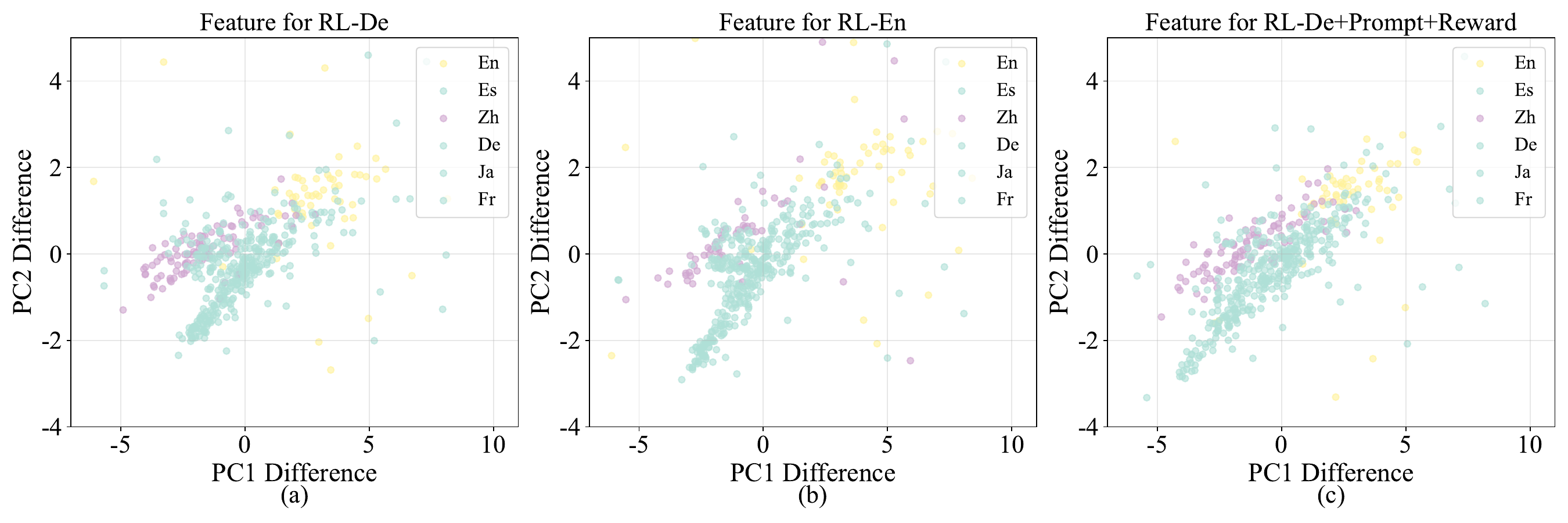}
     %%\vspace{-10pt}
    \caption{%
    Feature of LLM's hidden state of last layer, in training (dataset LUFFY) configuration of  (a)RL-De, (b)RL-De+Prompt, and (c))RL-De+Prompt+Reward. “+Prompt” adds language control prompts, and “+Reward” adds a language consistency reward.
    }
    \label{fig:RL-de}
\vspace{-10pt}
\end{figure}

\subsection{Exploration of model Semantic Feature shift}
 % @yiran
 % 为了进一步探究为什么RL不同语言训练会对结果造成影响，有些语言泛化性强 有些泛化性弱 我们进行了如下探究和分析
% To further investigate why RL training with different languages yields varying generalization capabilities, we conduct a semantic feature analysis of the learned representations. We analyze the hidden states of models trained under different configurations to understand the relationship between representation shift and cross-lingual generalization performance.
To investigate why RL training with different languages yields varying generalization capabilities, we analyze the semantic feature shifts in learned representations.

%  这个图表示了这三个配置 在LUFFY下训练得到的LLM 
% 这里对base模型，和这些模型 进行表征语义的测试 把6种语言MMATH500测试数据取来 计算最后一层的hidden state 然后通过PCA降维后得到数据$h_{Base}=\{(x^{Base}_1, y^{Base}_1, \cdots \}$ $h_{RL}=\{(x^{RL}_1, y^{RL}_1, \cdots \}$,取两者的差$h_{diff}=\{(x^{RL}_1-x^{Base}_1, y^{RL}_1-y^{Base}_1) \cdots \}$ 绘制散点图 
% \textbf{Methodology.} 
% We extract the final layer hidden states from the base model and RL-trained models when processing the MMath500 test data across six languages. These high-dimensional representations are projected to 2D space using Principal Component Analysis (PCA). For each configuration, we compute the difference vectors between the RL model and base model representations: $\mathbf{h}_{diff} = \mathbf{h}_{RL} - \mathbf{h}_{Base}$, where $\mathbf{h}_{RL}$
% and $\mathbf{h}_{Base}$ represent the 2D projections of the respective models' hidden states.
\textbf{Methodology.} We extract final layer hidden states from base and RL-trained models when processing MMath500 test data across six languages. These representations are projected to 2D space using PCA, and we compute difference vectors: $\mathbf{h}_{diff} = \mathbf{h}_{RL} - \mathbf{h}_{Base}$.

\textbf{Results.} Figure~\ref{fig:RL-de} reveals distinct shift patterns across training configurations. RL-De exhibits the most concentrated distribution around the origin, indicating minimal deviation from base representations, while RL-En displays more scattered distributions. This ordering directly correlates with cross-lingual performance in Table~\ref{tab:main table2}. Similarly, language consistency interventions in RL-De progressively increase representational scatter: baseline RL-De maintains compact distributions, while RL-De+Prompt+Reward shows greater dispersion, mirroring the performance degradation pattern.

%% 一种可能的解释是 Base 模型已经在pretrain 过程中建立了许多高层次的模式结构 例如回溯\citep{} 分支\citep{}。。。 后训练阶段 对这些结构的破坏越少 模型保留的有用模式就越多 泛化能力就越强\citep{}。对这种结构的破坏体现在表征偏移上，所以表征偏移较少的模型多语言泛化能力强。
%
% \textbf{Interpretation. }
% These findings suggest that the base model’s pre-training establishes high-level structural patterns and reasoning modules that are crucial for cross-lingual transfer \citep{hua2024mOthello}. 
% Models that preserve these structures—evidenced by minimal representational drift—maintain stronger generalization capabilities \citep{merchant2020what}. 
% The disruption of these pre-trained patterns, reflected in larger representational shifts, appears to impair the model’s ability to leverage language-agnostic reasoning mechanisms \citep{luo2023catastrophic}. 
% This provides a representational-level explanation for why RL’s linguistic inconsistency enhances rather than hinders cross-lingual performance: by avoiding forced adherence to specific languages, RL preserves the universal reasoning structures learned during pre-training \citep{lai2025rlforget}.
\textbf{Interpretation.} These findings suggest that pre-training establishes multilingual reasoning structures crucial for cross-lingual transfer \citep{hua2024mOthello,merchant2020what}. Models preserving these structures through minimal representational drift maintain stronger generalization capabilities. Conversely, larger shifts disrupt universal reasoning mechanisms \citep{luo2023catastrophic}, explaining why RL's linguistic inconsistency paradoxically enhances cross-lingual performance by preserving pre-trained reasoning structures \citep{lai2025rlforget}.
\section{Conclusion}

% 本文系统性地研究了强化学习（RL）与监督微调（SFT）在增强跨语言推理能力上的差异。我们的工作贡献了两个核心发现：首先，我们证明了RL不仅比SFT准确率更高，还展现出更强的跨语言泛化能力。其次，我们揭示了一个与传统认知相反的现象：使用非英语数据进行RL训练能带来更优越的性能，从而挑战了以英语为中心的训练范式。我们将这些优势归因于RL利用语言无关推理过程（“语言不一致性”）的能力及其固有的“探索-优化”机制。这些发现为构建更公平、高效的多语言推理系统提供了关键见解，为实现超越语言障碍的通用人工智能推理铺平了道路。
We systematically investigated the differences between Reinforcement Learning and Supervised Fine-Tuning for enhancing cross-lingual reasoning and the generalization across languages.
Multiple experiments demonstrate that RL not only achieves substantially higher accuracy than SFT but also exhibits superior cross-lingual generalization. Contrary to conventional cognition, we find that RL training on non-English data yields superior performance, challenging English-centric training. 
% Surprisingly, RL training on non-English data consistently outperforms English-based training—a phenomenon absent in SFT.
% Our work makes two central contributions: first, we demonstrate that RL not only achieves higher accuracy than SFT but also exhibits significantly stronger cross-lingual generalization. Second, we reveal that, contrary to conventional wisdom, RL training on non-English data yields superior performance, challenging the English-centric training paradigm. 
% Comprehensive mechanistic analysis reveals that RL models spontaneously break free from training language constraints, adopting inconsistent languages during reasoning, improving the generalization across languages. Furthermore, explore-and-optimize sampling strategy of RL also enables learning of robust and language-agnostic reasoning patterns. 
Our preliminary mechanistic analysis investigates the potential reasons for the superior cross-lingual generalization of RL from three perspectives: the linguistic inconsistency during the reasoning process, the unique explore-and-optimize sampling strategy, and the semantic shift after training. The understanding of these potential factors not only provides crucial insights into understanding RL's advantages in multilingual reasoning but also establishes a foundation for effectively enhancing cross-lingual reasoning in the future.

% We attribute these advantages to RL's ability to leverage a language-agnostic reasoning process ("linguistic inconsistency") and its inherent "explore-and-optimize" mechanism.
% These findings offer crucial insights for more equitable and effective multilingual reasoning.



%%%%%%%%%%%%%%%%%%%%%%%%%%%%%%%%%%%%%%%%%%%%%%%%%%%%%%%%%%%%%%%%
%%%%%%%%%%%%%%%%%%%%%%%% END OF MAIN %%%%%%%%%%%%%%%%%%%%%%%%
%%%%%%%%%%%%%%%%%%%%%%%%%%%%%%%%%%%%%%%%%%%%%%%%%%%%%%%%%%%%%%%%

% 不计入论文页数
% \section{Reproducibility statement}
% We ensure reproducibility by providing the multilingual dataset used in our main experiments as supplementary material, together with the training scripts for both RL and SFT frameworks and the core prompts. Detailed training settings and hyperparameters are reported in Section~\ref{sec:exp}, while preprocessing steps and additional implementation details are included in the Appendix~\ref{apd:exp}. 
% We also provide some \verb|requirements.txt| files to specify the training and evaluation environments.

\section{Ethics statement}
    
This study acknowledges several ethical implications of its investigation into cross-lingual reasoning in LLMs. Data and fairness concerns arise from potential biases in multilingual benchmarks, which may introduce performance disparities across languages. Our evaluations incorporate diverse linguistic settings and different multilingual reasoning tasks to mitigate such biases, though future work must further scrutinize cultural and linguistic influences on model behavior. 
% Additionally, research integrity is ensured through transparent methodologies, reproducible experiments using established benchmarks, and a commitment to public code release where permitted by licensing constraints (no proprietary data or human subjects were involved).

Beyond technical limitations, societal impact requires careful consideration. While improved multilingual reasoning could enhance accessibility for non-English speakers, reducing barriers in education and professional settings, it also risks misuse—such as automated disinformation generation or harmful content propagation across languages. We advocate for responsible deployment, emphasizing robust safeguards, human oversight, and ongoing risk assessments to balance innovation with ethical constraints.
%%%%%%%%%%%%%%%%%%%%%%%%%%%%%%%%%%%%%%%%%%%%%%%%%%%%%%%%%%%%%%%%
%%%%%%%%%%%%%%%%%%%%%%%% END OF CONTENT %%%%%%%%%%%%%%%%%%%%%%%%
%%%%%%%%%%%%%%%%%%%%%%%%%%%%%%%%%%%%%%%%%%%%%%%%%%%%%%%%%%%%%%%%

\bibliography{iclr2026_conference}
\bibliographystyle{iclr2026_conference}

\appendix
\section{Appendix}
\subsection{Use of LLM}
During the preparation of this work, large language models (e.g., ChatGPT) were used for English writing refinement and minor assistance in code debugging. All ideas, experiments, and analyses are solely by the authors.

\subsection{Experimental Settings}
\label{apd:exp}
\subsubsection{Prompts for generating translated GSM8K training datasets}
Here is the prompts for generating translated GSM8K training datasets.
\small{
\begin{verbatim}
TRANSLATION_PROMPT_TEMPLATE = """You are a professional math
translation assistant. Please translate the following English 
math problem into {target_language}, maintaining the mathematical
expressions and formatting.

Requirements:
1. Maintain the format of the mathematical calculation process 
(e.g., <<48/2=24>>)
2. Maintain the format of the final answer (e.g., #### 72)
3. The translation should be accurate and natural.
4. Keep the numbers and mathematical symbols unchanged.

Original question:

Original answer:

Please translate the question and answer separately, 
using the following format:
{{
"translated_question": "Translated question",
"translated_answer": "Translated answer"
}}

Please return the result in JSON format, using the 
{target_language} language, and do not add any additional text.
"""
\end{verbatim}
}
\subsubsection{Prompts for generating translated LUFFY training datasets}
Here is the prompts for generating translated LUFFY training datasets.
\small{
\begin{verbatim}
TRANSLATION_PROMPT_TEMPLATE = """You are a professional math 
translation assistant. Please translate the following content
into {target_language}, preserving mathematical expressions,
LaTeX formulas, and special formatting.

Requirements:
1. Keep all mathematical formulas and LaTeX expressions
intact (e.g., $24 \\mathrm{{~km}}$, \\boxed{{}}, etc.)
2. Keep the <think> and </think> tags intact
3. The translation should be accurate and natural.
4. Keep numbers and mathematical symbols intact.

Original content:
{content}

Please return the translated {target_language} content
directly in the format
{{
"translated_content": "Translated content"
}}
Do not add any additional explanatory text.

"""
\end{verbatim}
}

"\subsection{Results details}
\label{apd:results}

\begin{table*}[h]
 \centering
\caption{Performance of base model, SFT, and RL tuning models on MGSM. Base denotes the orginal Qwen2.5-3B-Base model. SFT (zh) and RL (zh) mean we tune the base model in Chinese data through SFT and RL, respectively. We report the accuracy score on 10 linguistic settings. $\Delta$ (RL-SFT) represents the performance difference between RL and the corresponding SFT score. Each score represents the average accuracy over six measurements. Avg represents the average of the scores of 10 language settings and Gen represents the generalization score.}
 \begin{adjustbox}{width=.98\textwidth}{
\begin{tabular}{lcccccccccccc} 
\toprule
Models            & En & Zh     & De & Es & Fr     & Ja & Ru & Th & Sw         & Bn          & Avg         & Gen     \\ 
\midrule
Base & 63.4~           & 48.3~  & 33.5~           & 57.7~           & 38.9~  & 19.5~           & 30.3~           & 17.6~           & 7.3~       & 1.2~        & 31.8~       & 0.0~    \\ 
\midrule
SFT (En)          & 64.7~           & 54.5~  & 50.7~           & 56.4~           & 56.2~  & 36.9~           & 55.5~           & 44.1~           & 6.9~       & 26.2~       & 45.2~       & 18.1~   \\
RL (En)           & 85.8~           & 72.1~  & 70.8~           & 77.3~           & 76.6~  & 61.2~           & 64.9~           & 61.0~           & 9.5~       & 47.5~       & 62.7~       & 49.1~   \\
$\Delta$ (RL-SFT) & +21.1~          & +17.6~ & +20.1~          & +20.9~          & +20.4~ & \textbf{+24.3}~ & +9.4~           & +16.9~          & +2.6~      & \textbf{\textit{+21.3}}~ & +17.5~      & +30.9~  \\ 
\midrule
SFT (Zh)          & 65.7~           & 58.7~  & 48.4~           & 55.7~           & 56.1~  & 43.5~           & 56.6~           & 45.8~           & 7.5~       & 30.5~       & 46.9~       & 20.4~   \\
RL (Zh)           & 86.1~           & 76.3~  & 74.2~           & 81.1~           & 76.1~  & 64.5~           & 78.1~           & 64.9~           & 10.3~      & 48.3~       & 66.0~       & 52.6~   \\
$\Delta$ (RL-SFT) & +20.4~          & +17.6~ & \textbf{+25.8}~ & \textbf{+25.4}~ & +20.0~ & +21.0~          & +21.5~          & +19.1~          & +2.8~      & +17.8~      & +19.1~      & +32.3~  \\ 
\midrule
SFT (De)          & 63.9~           & 54.2~  & 57.5~           & 55.7~           & 52.8~  & 39.3~           & 55.1~           & 47.6~           & 8.4~       & 28.8~       & 46.3~       & 19.3~   \\
RL (De)           & 91.0~           & 77.6~  & 80.5~           & 82.7~           & 80.0~  & 67.8~           & 81.3~           & 75.3~           & 15.9~      & 63.3~       & 71.5~       & 60.4~   \\
$\Delta$ (RL-SFT) & \textit{+27.1}~ & +23.4~ & +23.0~          & +27.0~          & +27.2~ & +28.5~          & +26.2~          & +27.7~          & \textbf{\textit{+7.5}}~ & \textbf{+34.5}~          & \textbf{\textit{+25.2}}~ & \textbf{\textit{+41.2}}~  \\ 
\midrule
SFT (Es)          & 63.9~           & 54.7~  & 54.3~           & 62.7~           & 54.0~  & 41.1~           & 58.1~           & 46.5~           & 9.5~       & 31.1~       & 47.6~       & 21.6~   \\
RL (Es)           & 89.3~           & 77.8~  & 78.0~           & 82.1~           & 77.3~  & 68.9~           & 80.3~           & 72.7~           & 13.4~      & 53.7~       & 69.4~       & 57.5~   \\
$\Delta$ (RL-SFT) & +25.4~          & +23.1~ & +23.7~          & +19.4~          & +23.3~ & +27.8~          & +22.2~          & +26.2~          & +3.9~      & +22.6~      & +21.8~      & +35.9~  \\ 
\midrule
SFT (Fr)          & 64.8~           & 53.7~  & 51.3~           & 58.8~           & 57.9~  & 40.9~           & 57.1~           & 46.5~           & 8.9~       & 29.9~       & 47.0~       & 20.6~   \\
RL (Fr)           & 89.3~           & 78.9~  & 77.5~           & 82.3~           & 81.1~  & 70.9~           & 81.1~           & 73.1~           & 13.3~      & 59.1~       & 70.7~       & 59.3~   \\
$\Delta$ (RL-SFT) & +24.5~          & +25.2~ & +26.2~          & +23.5~          & +23.2~ & \textbf{+30.0}~ & +24.0~          & +26.6~          & +4.4~      & +29.2~      & +23.7~      & +38.7~  \\ 
\midrule
SFT (Ja)          & 64.4~           & 56.5~  & 50.5~           & 58.2~           & 53.6~  & 51.3~           & 54.3~           & 45.6~           & 8.1~       & 33.7~       & 47.6~       & 21.1~   \\
RL (Ja)           & 88.1~           & 79.1~  & 78.8~           & 81.5~           & 79.3~  & 72.7~           & 81.7~           & 72.4~           & 14.1~      & 61.7~       & 70.9~       & 59.2~   \\
$\Delta$ (RL-SFT) & +23.7~          & +22.6~ & +28.3~          & +23.3~          & +25.7~ & +21.4~          & \textbf{+27.4}~ & +26.8~          & +6.0~      & +28.0~      & +23.3~      & +38.1~  \\ 
\midrule
SFT (Ru)          & 64.8~           & 54.9~  & 53.5~           & 56.7~           & 55.1~  & 39.5~           & 57.3~           & 44.9~           & 10.4~      & 29.8~       & 46.7~       & 20.0~   \\
RL (Ru)           & 87.5~           & 76.6~  & 78.5~           & 79.9~           & 78.8~  & 69.6~           & 80.3~           & 73.5~           & 12.5~      & 57.8~       & 69.5~       & 57.1~   \\
$\Delta$ (RL-SFT) & +22.7~          & +21.7~ & +25.0~          & +23.2~          & +23.7~ & +30.1~          & +23.0~          & \textbf{+28.6}~ & +2.1~      & +28.0~      & +22.8~      & +37.1~  \\
\bottomrule
\end{tabular}
}
\end{adjustbox}
\label{tab:main table1 apd}
\end{table*}

\begin{table}[h]
\centering
% \small
\caption{Performance of base model, SFT, and RL tuning models on MAIME2024. Base denotes the orginal Qwen2.5-3B-Base model. SFT (zh) and RL (zh) mean we tune the base model in Chinese data through SFT and RL, respectively. We report the Pass@16 score on 10 linguistic settings. $\Delta$ (RL-SFT) represents the performance difference between RL and the corresponding SFT score.}
\label{tab:appendix table1}
 \begin{adjustbox}{width=.98\textwidth}{
\begin{tabular}{lccccccccccc} 
\toprule
Models            & Zh     & Fr     & En     & De     & Ja     & Es     & Ru     & Th     & Bn     & Sw     & Average  \\ 
\midrule
Base              & 6.7~   & 6.7~   & 16.7~  & 10.0~  & 10.0~  & 13.3~  & 3.3~   & 10.0~  & 3.3~   & 0.0~   & 8.0~     \\ 
\midrule
SFT (Zh)          & 20.0~  & 6.7~   & 13.3~  & 6.7~   & 10.0~  & 13.3~  & 10.0~  & 20.0~  & 6.7~   & 3.3~   & 11.0~    \\
RL (Zh)           & 26.7~  & 30.0~  & 23.3~  & 26.7~  & 26.7~  & 23.3~  & 23.3~  & 26.7~  & 20.0~  & 10.0~  & 23.7~    \\
$\Delta$ (RL-SFT) & +6.7~  & +23.3~ & +10.0~ & +20.0~ & +16.7~ & +10.0~ & +13.3~ & +6.7~  & +13.3~ & +6.7~  & +12.7~   \\ 
\midrule
SFT (De)          & 13.3~  & 13.3~  & 16.7~  & 13.3~  & 0.0~   & 6.7~   & 10.0~  & 3.3~   & 6.7~   & 6.7~   & 9.0~     \\
RL (De)           & 26.7~  & 20.0~  & 20.0~  & 23.3~  & 13.3~  & 30.0~  & 26.7~  & 20.0~  & 13.3~  & 16.7~  & 21.0~    \\
$\Delta$ (RL-SFT) & +13.4~ & +6.7~  & +3.3~  & +10.0~ & +13.3~ & +23.3~ & +16.7~ & +16.7~ & +6.6~  & +10.0~ & +12.0~   \\
\bottomrule
\end{tabular}}
    \end{adjustbox}
\end{table}

% \clearpage
% 
\begin{figure*}[h]
    \centering
    \small
    
    % \subfloat[A comparison of performance on MGSM.]{
    %     \includegraphics[width=0.8\textwidth]{gsm8k.pdf}
    %     \label{fig:bar1}
    % }
    
    % %%\vspace{0.5cm} % 调整子图间的垂直间距
    
    \subfloat[A comparison of performance on MMath500.]{
        \includegraphics[width=0.75\textwidth]{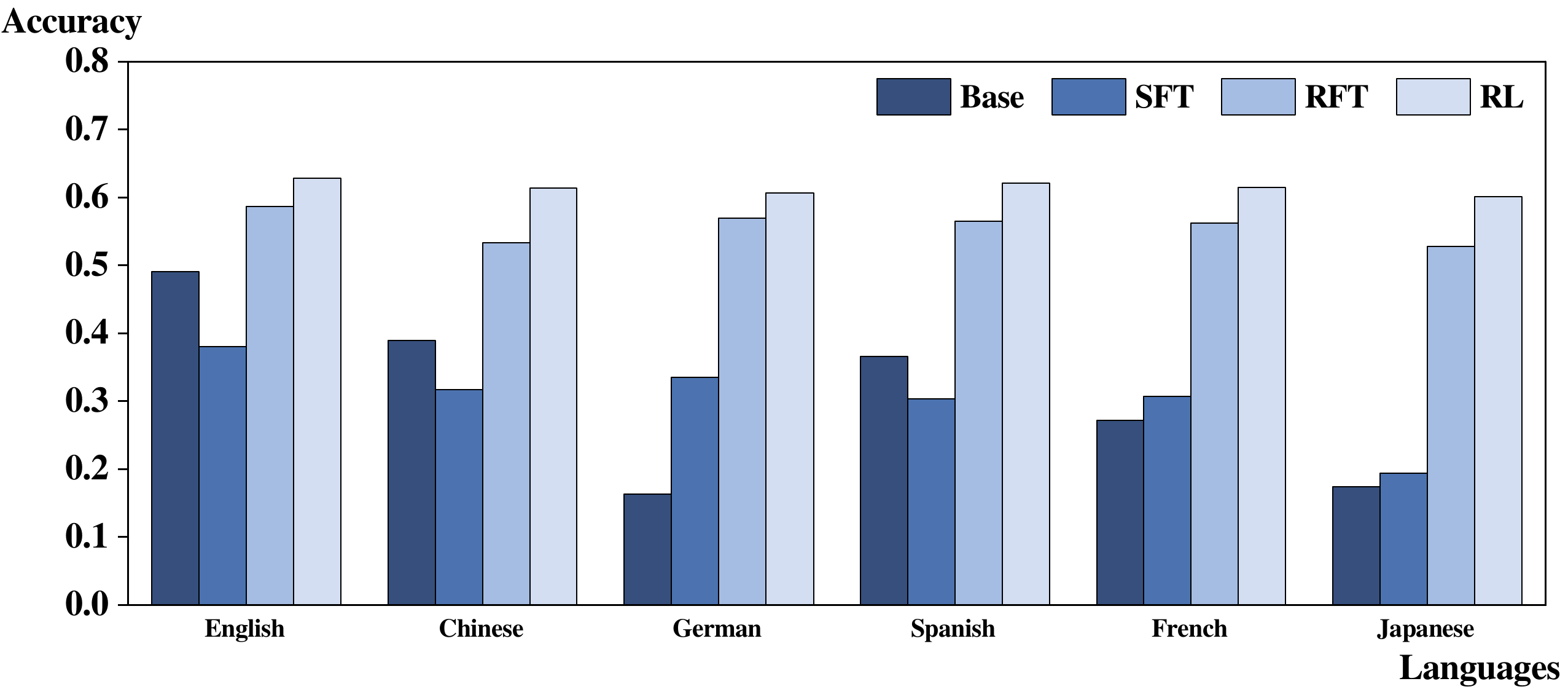}
        \label{fig:bar2}
    }
    
    % %%\vspace{0.5cm} % 调整子图间的垂直间距
    
    \subfloat[A comparison of performance on MAIME2024.]{
        \includegraphics[width=0.75\textwidth]{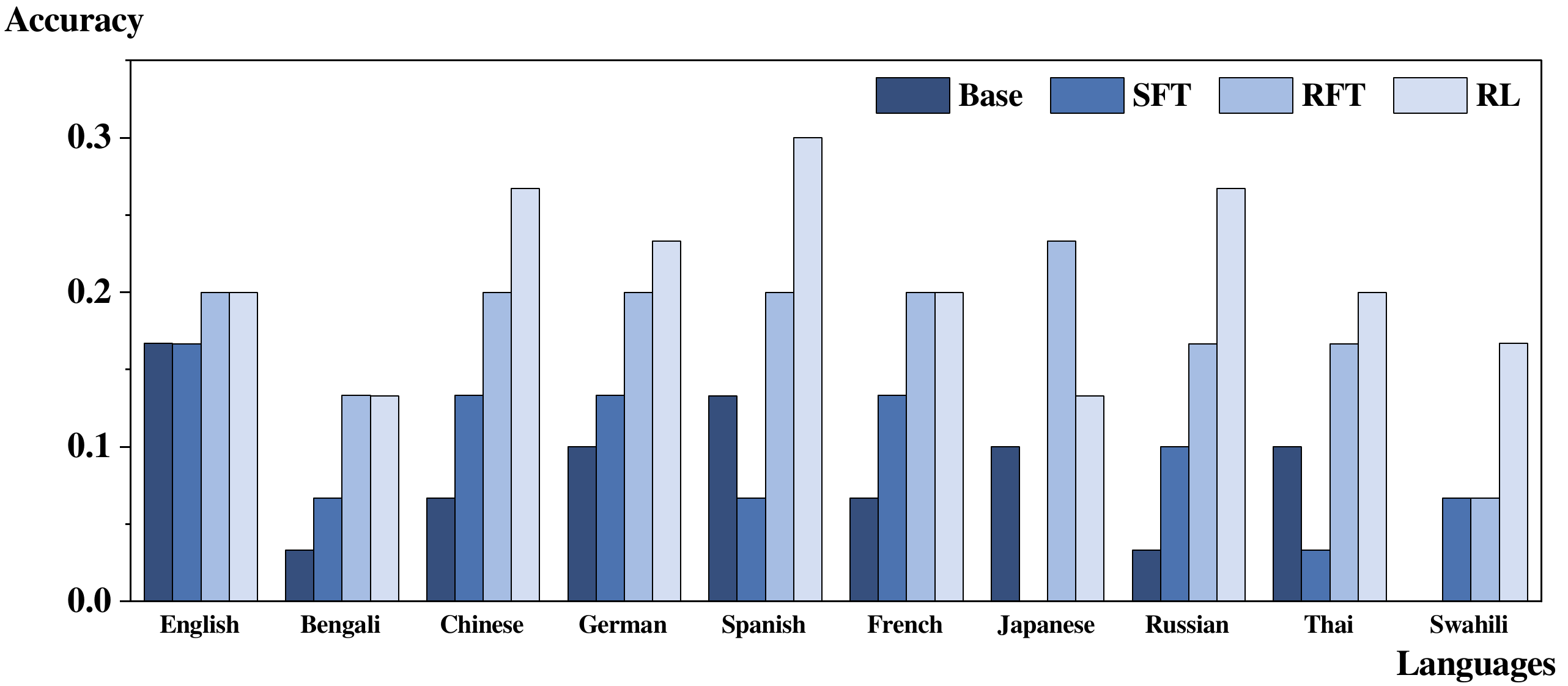}
        \label{fig:bar3}
    }
    
    \caption{Model performance comparisons among the Base, SFT, RFT, and RL models. We use German data in LUFFY in SFT, RL, RFT for training.}
    \label{fig:combined-apd}
\end{figure*}

\clearpage
\begin{table}[h]
\centering
\caption{Performance of models on MMath500. 
% We report Base denotes the orginal Qwen2.5-3B-Base model. 
% RL (zh), (+ Prompt) and (+ Consistency Reward) mean the model trained with RL in Chinese data, adding language control in prompt, further adding language consistency reward in reward, respectively. 
``RL (zh)'' denotes the model trained with Reinforcement Learning on Chinese data. ``+ Prompt'' indicates the addition of language control prompts during both the training and the inference. ``+ Consistency Reward'' further incorporates a language consistency reward into the training objective.
We report the accuracy score on 6 linguistic settings. We test 6 times and report the average accuracy scores and pass@k scores.
}
\label{tab:control language}
\begin{adjustbox}{width=.8\textwidth}{
\begin{tabular}{lccccccc} 
\toprule
Models     & Zh   & Fr   & En   & De   & Ja   & Es   & ~ Average  \\ 
\midrule
\multicolumn{8}{c}{\textbf{‌Average Scores}}         \\ 
\midrule
Base                          & 38.9 & 27.2 & 49.1 & 16.3 & 17.4 & 36.6 & 30.9       \\ 
RL (En)                       & 53.7 & 55.8 & 62.7 & 50.9 & 54.2 & 56.6 & 55.7       \\
\midrule
RL (Zh)                       & 61.3 & 61.2 & 63.3 & 61.2 & 58.5 & 62.1 & 61.3       \\
+ Prompt                      & 53.3 & 54.9 & 59.7 & 42.2 & 55.1 & 56.8 & 53.7       \\
+ Prompt + Consistency Reward & 56.2 & 54.2 & 62.9 & 45.5 & 48.2 & 56.4 & 53.9       \\ 
\midrule
RL (De)                       & 61.4 & 61.5 & 62.8 & 60.7 & 60.1 & 62.1 & 61.4       \\
+ Prompt                      & 56.0   & 61.3 & 63.8 & 60.8 & 59.0   & 61.9 & 60.5       \\
+ Prompt + Consistency Reward & 51.9 & 52.1 & 62.4 & 49.3 & 41.6 & 54.6 & 52.0         \\
\midrule
\multicolumn{8}{c}{\textbf{Pass@6 Scores}}         \\ 
\midrule
Base                          & 67.9 & 60.9 & 75.8 & 48.1 & 44.9 & 69.3 & 61.2       \\
RL (En)                       & 74.7 & 77.2 & 78.8 & 73.9 & 75.4 & 77.6 & 76.3       \\ 
\midrule
RL (Zh)                       & 78.4 & 76.2 & 81.4 & 78.0   & 75.2 & 77.8 & 77.8       \\
+ Prompt                      & 73.9 & 76.8 & 77.0   & 72.7 & 74.5 & 77.2 & 75.4       \\
+ Prompt + Consistency Reward & 74.1 & 73.7 & 81.4 & 72.7 & 70.9 & 76.2 & 74.8       \\ 
\midrule
RL (De)                       & 78.4 & 78.8 & 79.0   & 76.4 & 77.8 & 78.8 & 78.2       \\
+ Prompt                      & 77.8 & 79.4 & 80.4 & 77.8 & 77.8 & 79.2 & 78.7       \\
+ Prompt + Consistency Reward & 74.1 & 73.9 & 79.0   & 72.9 & 65.3 & 76.2 & 73.6       \\
\bottomrule
\end{tabular}
}\end{adjustbox}
\end{table}

To complement Figure~\ref{fig:RL-de}, Table~\ref{tab:model_PCA} reports the quantitative measurements of representational movement under different RL configurations. Specifically, ``Model Center Distance'' denotes the distance between each model’s representation center and the base model center, while ``Model Shift Distance'' denotes the distance between the model’s shift center and the zero point. These measurements provide quantitative evidence supporting the representational patterns illustrated in the figure.

\begin{table}[h]
\centering
\caption{Numerical results corresponding to Figure~\ref{fig:RL-de}, reporting the model center distance and shift distance under different RL configurations.}
\label{tab:model_PCA}
\begin{tabular}{lcc}
\toprule
Config              & Model Center Distance & Model Shift Distance \\
\midrule
RL-En               & 2.255                 & 41.332               \\
RL-Zh               & 1.815                 & 41.294               \\
RL-De               & 1.753                 & 41.241               \\
RL-De+Prompt        & 1.891                 & 41.286               \\
RL-De+Prompt+Reward & 1.908                 & 41.652             \\
\bottomrule
\end{tabular}
\end{table}

\clearpage
\subsection{Case Study}
\label{apd:case}
% \begin{CJK*}{UTF8}{gbsn}
\begin{table*} [h]
\centering
\caption{Case study on MMath500. We select the German question to show the case.}
 %%\vspace{-8pt}
\label{tab:samples-apd}
\begin{adjustbox}{width=.9\textwidth}{
\begin{tabular}{l|l} 
\toprule
    \multicolumn{2}{l}{\begin{tabular}[c]
    {@{}p{0.96\linewidth}@{}}
    \sethlcolor{blue!10}\hl{\textbf{Question:}} Ein Matheclub veranstaltet einen Kuchenverkauf als Spendenaktion, um Geld für eine bevorstehende Reise zu sammeln. Sie verkaufen $54$ Kekse zu drei für \$1\$, und \$20\$ Cupcakes zu \$2\$ pro Stück, und \$35\$ Brownies zu \$1\$ pro Stück. Wenn es den Matheclub \$15\$ gekostet hat, diese Artikel zu backen, wie hoch war ihr Gewinn? \\
    \textbf{Answer}: 78\\
    \end{tabular}} \\ 
\midrule
    \multicolumn{2}{l}{\begin{tabular}[c]{@{}p{0.96\linewidth}@{}}
    \sethlcolor{blue!10}\hl{\textbf{Base:}}\\
    % \textbf{Original Awareness Information}: 
    Da Kekse zu drei für $\$1$ verkaufen werden, müssen wir zahlenweise Kekse zählen. Es gibt $54$ Kekse zu backen, und jeder Kuchen besteht aus drei Keksen. Also haben sie $54/3=18$ Kuchen zu backen. Zudem haben sie $20$ Cupcakes zu backen und $35$ Brownies zu backen. Die Matheclub hat also insgesamt $18+20+35=73$ Artikel zu backen. Sind sie zu backen, hat es ihnen $73 \times 2 = \$146$ gekostet. Da sie für die Backzeit $15$ Dollar ausgeben, haben sie für den Verkauf $146-15 = \$131$ eingenommen. Daher haben sie einen Gewinn von $\boxed{131}$.
    \end{tabular}} \\ 
\midrule
    \multicolumn{2}{l}{\begin{tabular}[c]{@{}p{0.96\linewidth}@{}}
    \sethlcolor{blue!10}\hl{\textbf{Reinforcement Learning (with German Data):}}\\
    % \textbf{Original Awareness Information}: 
    
    1. Calculate the revenue from Kekse:
       - They sold 54 Kekse at a rate of 3 Kekse for \$1.
       - The number of sets of 3 Kekse in 54 is \( \frac{54}{3} = 18 \).
       - Revenue from Kekse = \( 18 \times 1 = 18 \) dollars.
    
    2. Calculate the revenue from Cupcakes:
       - They sold 20 Cupcakes at \$2 each.
       - Revenue from Cupcakes = \( 20 \times 2 = 40 \) dollars.
    
    3. Calculate the revenue from Brownies:
       - They sold 35 Brownies at \$1 each.
       - Revenue from Brownies = \( 35 \times 1 = 35 \) dollars.
    
    4. Calculate the total revenue:
       - Total revenue = Revenue from Kekse + Revenue from Cupcakes + Revenue from Brownies
       - Total revenue = \( 18 + 40 + 35 = 93 \) dollars.
    
    5. Calculate the profit:
       - Cost to make the items = \$15.
       - Profit = Total revenue - Cost to make the items
       - Profit = \( 93 - 15 = 78 \) dollars.
    
    Therefore, the profit is \(\boxed{78}\).
    \end{tabular}} \\ 
\midrule
    \multicolumn{2}{l}{\begin{tabular}[c]{@{}p{0.96\linewidth}@{}}
    \sethlcolor{blue!10}\hl{\textbf{Reinforcement Learning (with German Data and with Consistency Reward):}}\\
    Zuerst berechnen wir den Einnahmen von den Keksen. Da sie drei Kekse für $\$1$ verkaufen, erhalten sie:
    $54 \div 3 = 18 \text{ Doppelsätze}$, $18 \times 2 = 36 \text{ Dollar}$
    Für die Cupcakes erhalten sie:
    $20 \div 2 = 10 \text{ Stück}$, $10 \times 2 = 20 \text{ Dollar}$
    Für die Brownies erhalten sie:
    $35 \div 1 = 35 \text{ Stück}$, $35 \times 1 = 35 \text{ Dollar}$
    Die Gesamtsumme ihrer Einnahmen beträgt:
    $36 + 20 + 35 = 91 \text{ Dollar}$
    Da es ihnen $\$15$ gekostet hat, um die Artikel zu backen, erhalten sie:
    $91 - 15 = 76 \text{ Dollar}$
    Die Gewinnsumme beträgt \boxed{76}.
    \end{tabular}} \\
% \midrule
\bottomrule
\end{tabular}
}\end{adjustbox}
 %%\vspace{-10pt}
\end{table*}
% \end{CJK*}

\end{document}